\documentclass{article}

\PassOptionsToPackage{numbers, compress}{natbib}
\usepackage[preprint, nonatbib]{neurips_2024}

\usepackage[utf8]{inputenc} %
\usepackage[T1]{fontenc}    %
\usepackage{hyperref}       %
\usepackage{url}            %
\usepackage{booktabs}       %
\usepackage{amsfonts}       %
\usepackage{nicefrac}       %
\usepackage{microtype}      %
\usepackage{xcolor}         %

\usepackage{sidecap} 
\usepackage{graphicx} 
\usepackage{epsfig}
\usepackage{amsmath}
\usepackage{amssymb}
\usepackage{adjustbox}
\usepackage{amsmath}

\newcommand{\linkk}[1]{\emph{\textcolor{magenta}{#1}}}

\newcommand\nnfootnote[1]{%
  \begin{NoHyper}
  \renewcommand\thefootnote{}\footnote{#1}%
  \addtocounter{footnote}{-1}%
  \end{NoHyper}
}

\newcommand\ourmethod{Lay-A-Scene}
\newcommand\ourpnp{SI-PnP}
\newcommand\sidepnp{Side-Information PnP}

\title{\ourmethod{}: Personalized 3D Object Arrangement Using Text-to-Image Priors}

\author{%
  Ohad Rahamim$^*$ \\
  Bar-Ilan University\\
  \And
  Hilit Segev$^*$ \\
  Bar-Ilan University\\
  \And
  Idan Achituve \\
  Bar-Ilan University\\
  \And
  Yuval Atzmon \\
  NVIDIA Research\\
  \And
  Yoni Kasten \\
  NVIDIA Research\\ 
  \And
  Gal Chechik \\
  NVIDIA Research\\
  Bar-Ilan University
}

\begin{document}

\maketitle
\nnfootnote{These authors contributed equally to this work}

\begin{figure}[ht]
    \centering
    \includegraphics[width=\textwidth]{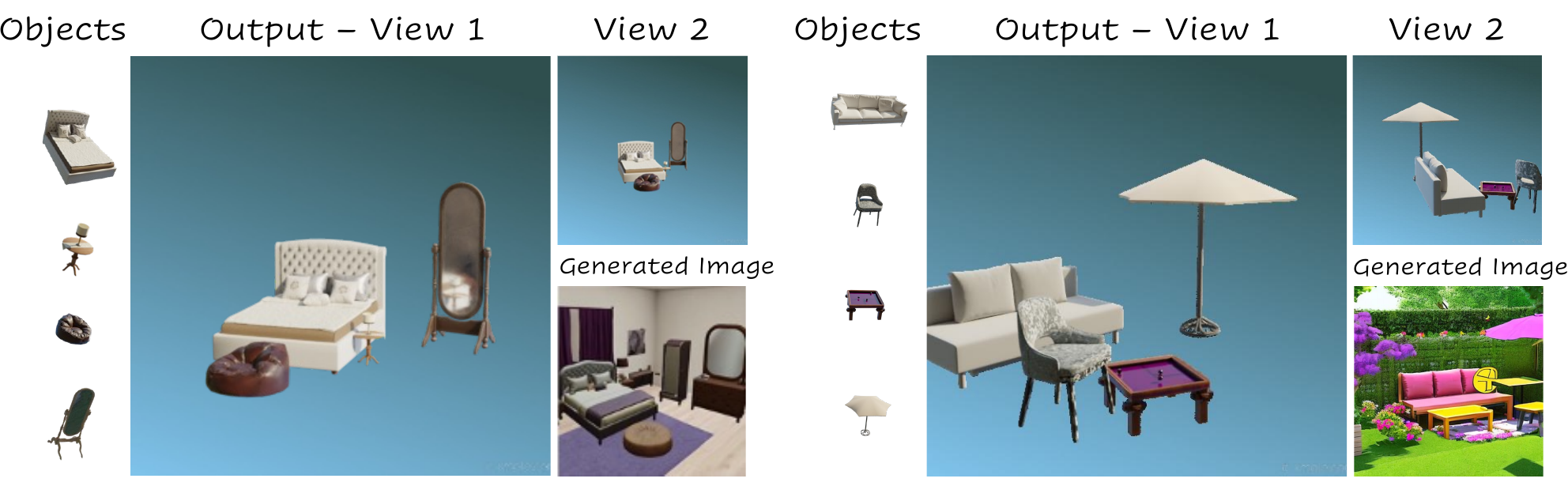}
    \caption{\ourmethod{}  is a test-time optimization method for finding plausible layouts of given 3D objects, leveraging a pre-trained text-to-image diffusion model. Given as input several mesh objects (left in each panel) and a textual description of the scene (``a bedroom'' or ''garden''), the text-to-image model generates a scene image with these objects (on the bottom-right). This image is used to find an arrangement of the 3D input objects (at the center of each panel).} 
    \label{fig:teaser}
\end{figure}

\begin{abstract}
    Generating 3D visual scenes is at the forefront of visual generative AI, but current 3D generation techniques struggle with generating scenes with multiple high-resolution objects. Here we introduce \textit{\ourmethod{}}, which solves the task of \textit{Open-set 3D Object Arrangement}, effectively arranging unseen objects. Given a set of 3D objects, the task is to find a plausible arrangement of these objects in a scene. We address this task by leveraging pre-trained text-to-image models. 
    We personalize the model and explain how to generate images of a scene that contains multiple predefined objects without neglecting any of them.
    Then, we describe how to infer the 3D poses and arrangement of objects from a 2D generated image by finding a consistent projection of objects onto the 2D scene. We evaluate the quality of \ourmethod{} using 3D objects from Objaverse and human raters and find that it often generates coherent and feasible 3D object arrangements. \\
    Project page: \href{https://lay-a-scene.github.io/}{\linkk{https://lay-a-scene.github.io/}}
\end{abstract}

\section{Introduction}

\label{sec:intro}

Generating 3D scenes is currently at the forefront of visual generative AI. Just as large-scale text-to-image diffusion models have revolutionized the way we think about creating imaginative visual content from textual descriptions, generating 3D scenes is key for many applications. This includes home design, virtual reality, and game development, where the ability to rapidly prototype and visualize 3D scenes from text descriptions can revolutionize the design process.

Since 3D training datasets are much smaller than image datasets, 
a central idea in this field is to distill the knowledge about objects and scenes from text-to-image models to guide the generation of 3D objects and scenes, but such distillation is challenging. 
Distillation methods based on SDS  \cite{poole2022dreamfusion} may collapse to typical modes of the distribution, and tend to err when generating less-sampled views of objects (known as the Janus problem) \cite{raj2023dreambooth3d}. Also, these methods are tasked with solving two different problems simultaneously: Guide generation of every 3D object, and also guide generation of an arrangement of all objects into a plausible scene \cite{gan2023idesigner}. 

In many cases, however, one may have access to existing models of 3D objects and is only interested in finding plausible arrangements of these objects \cite{jiang2012learning}. This setup, which we call here \textit{Open-Set-3D-Arrange}, can be viewed as a 3D parallel of the personalization problem in image generation \cite{TextualInversion,ruiz2023dreambooth}. 
It can also be seen as complementary to scene-generation approaches like those in \cite{make-a-scene}, which generate objects in a scene \textit{given} a layout. 3D-Arrange is about the reverse problem -- finding a plausible layout for given 3D objects.  3D object arrangement has been previously addressed \cite{jiang2012learning} by training models using specific training datasets. 
The question remains whether this problem can be solved by distilling information from current text-to-image models.

Here we describe $\ourmethod{}$, a method for addressing the 3D-object-Arrangement task. The key idea, is that since the objects already exist, we can use the text-to-image models to generate a single image that serves as a layout for arranging the objects. Instead of performing several inference passes with those models to generate a full scene, we generate an image with a plausible layout using a single forward pass. Then, we match each object and its appearance in the generated image to infer its position. Together, those positions for all objects create a full scene.

Several key challenges emerge when taking this approach. 
First, how can we infer 3D position of given objects from generated images? We propose to first personalize the text-to-image model using a few renderings of the given objects. Second, the arrangement in the 2D image may be physically infeasible, like in the case where objects are distorted or colliding. We propose using Perspective-n-Points (PnP) \cite{lu2018review},  a robust procedure for matching keypoints of 3D objects with their depictions in the scene. Third, matching 3D positions with a 2D image is not well-defined, as there could be several 3D positions that match a given image (the "Pisa tower on hand palm" illusion).  To address this, we show how to add prior information about physical considerations as soft constraints to the PnP optimization to generate scenes that are more coherent and natural.
We call this approach \textit{Side-information-PNP}, and find that it greatly improves the generated scenes. 
We also note that text-to-image models often suffer from \textit{entity neglect}, where generated images do not contain all entities mentioned in their prompts \cite{chefer2023attendandexcite, rassin2023linguistic}. This problem becomes more severe when objects are unusual or incoherent, and when the number of objects increases. 
In our setting, the PnP procedure can be used to filter out images with entity neglect.  

The \ourmethod{} approach has several main advantages. Unlike other 3D-arrange methods \cite{jiang2012learning} it can handle new objects through personalization without retraining the foundation model. Unlike graph-based 3D arrangement methods like \cite{zhai2023commonscenes}, which expect users to provide a set of spatial relations, \ourmethod{} generates the scene layout based on the prior learned by text-to-image diffusion models.

In summary, this paper makes the following contributions: 
\textbf{(1)} A new learning setup, Open-set-3D-Arrange: generating a plausible layout for a given set of new 3D objects.
\textbf{(2)} A layout generation approach that efficiently finds plausible translations and rotations parameters for given objects. 
\textbf{(3)} A method to introduce side information into the perspective-n-point  optimization procedure in terms of soft constraints. 
\textbf{(4)} State-of-the-art 3D scene synthesis results from a dataset of diverse objects.

\section{Related Work}

\textbf{3D object Arrangement - generating a 3D scene layout.}
Most relevant to our approach are works that learned to predict spatial relations between objects.   
\cite{yu2011make, fisher2012example} learns a probabilistic model over spatial relationships between objects from data on the placement of furniture indoors. This method generalizes well to objects in their training set, but not outside the distribution. More recently, LegoNet \cite{wei2023lego} is a 
 transformer-based iterative method for learning regular rearrangement of Objects in messy rooms. 

\textbf{3D Scene Synthesis.}
In the field of 2D computer vision, controllable scene synthesis has been extensively explored.
This spans various approaches, ranging from generating scenes based on text \cite{gafni2022make} to tasks involving segmentation maps \cite{park2019semantic} or spatial layout-based image-to-image translation \cite{zhao2021luminous}. In the 3D domain, conventional methods frequently relied on the availability of a 3D database for synthesis \cite{funkhouser2004modeling, kreavoy2007model, chaudhuri2010data}, for instance, assembling new objects from existing object parts \cite{funkhouser2004modeling}. These methods often used non-parametric techniques, or used probabilistic graphical models \cite{kreavoy2007model, chaudhuri2010data, merrell2010computer, chaudhuri2011probabilistic, jiang2012learning, kalogerakis2012probabilistic, shen2012structure, xu2012fit, liu2014creating, huang2015single}. %

More recent methods use deep neural networks for synthesis tasks, creating scenes from images \cite{tulsiani2018factoring, park2019semantic, yang2021indoor, zhao2021luminous, bahmani2023cc3d}, scene grammars \cite{devaranjan2020meta, purkait2020sg}, and scene layouts \cite{wang2018deep, jyothi2019layoutvae, dhamo2021graph,tang2023diffuscene}. Noteworthy is CommonScenes \cite{zhai2023commonscenes} which generates 3D scenes based on a scene graph and prompt features and is trained in an end-to-end fashion. Another avenue for creating scenes is based on a textual description. Some studies \cite{jain2022zeroshot,lee2022understanding,mohammad2022clip,jiang20233d} use vision-language models such as CLIP \cite{radford2021learning}. Some also refine a 3D input \cite{michel2022text2mesh,chen2022tango,wang2023nerf,richardson2023texture}. More related to our approach are methods that use pre-trained diffusion models \cite{rombach2022high,saharia2022photorealistic} to generate 3D content without training \cite{poole2022dreamfusion, lin2023magic3d, metzer2023latent,melas2023realfusion,wang2023score}. Specifically, DiffuScene~\cite{tang2023diffuscene} designs a scene-graph diffusion model, but it requires a reference scene and can generate objects from a closed-set. Similarly, RoomDreamer~\cite{SongCXKTYY23} generates a novel room based on a pre-scanned image of a room and a textual prompt. And, Text2Room \cite{hollein2023text2room} generates a mesh in an iterative fashion through in-painting and monocular depth estimation. However, those methods lift 2D-generated images to point clouds and are not applied to given objects.
Unlike these approaches, our \ourmethod{} is designed to arrange a set of given 3D objects.

\section{Method}
\subsection{Problem setup and overview of our method} \label{method}
Our goal is to organize a given set of 3D objects into a plausible arrangement, forming a 3D scene. 
Formally, we are given a set of $N$ input objects $\{O_i \}_{i=1,...,N}$ and a text description of the scene $y$. Our objective is to discover transformations $\{Tr_i \}_{i=1,...,N}$ for each object, where $Tr_i\in\mathit{SE}(3)$ 
represents a transformation matrix. We assume we have access to a text-to-image model like \cite{StableDiffusion}.
\begin{figure}[t]
    \centering
    \includegraphics[width=\textwidth, trim={2.cm 0.cm 0.cm 0.cm},clip]{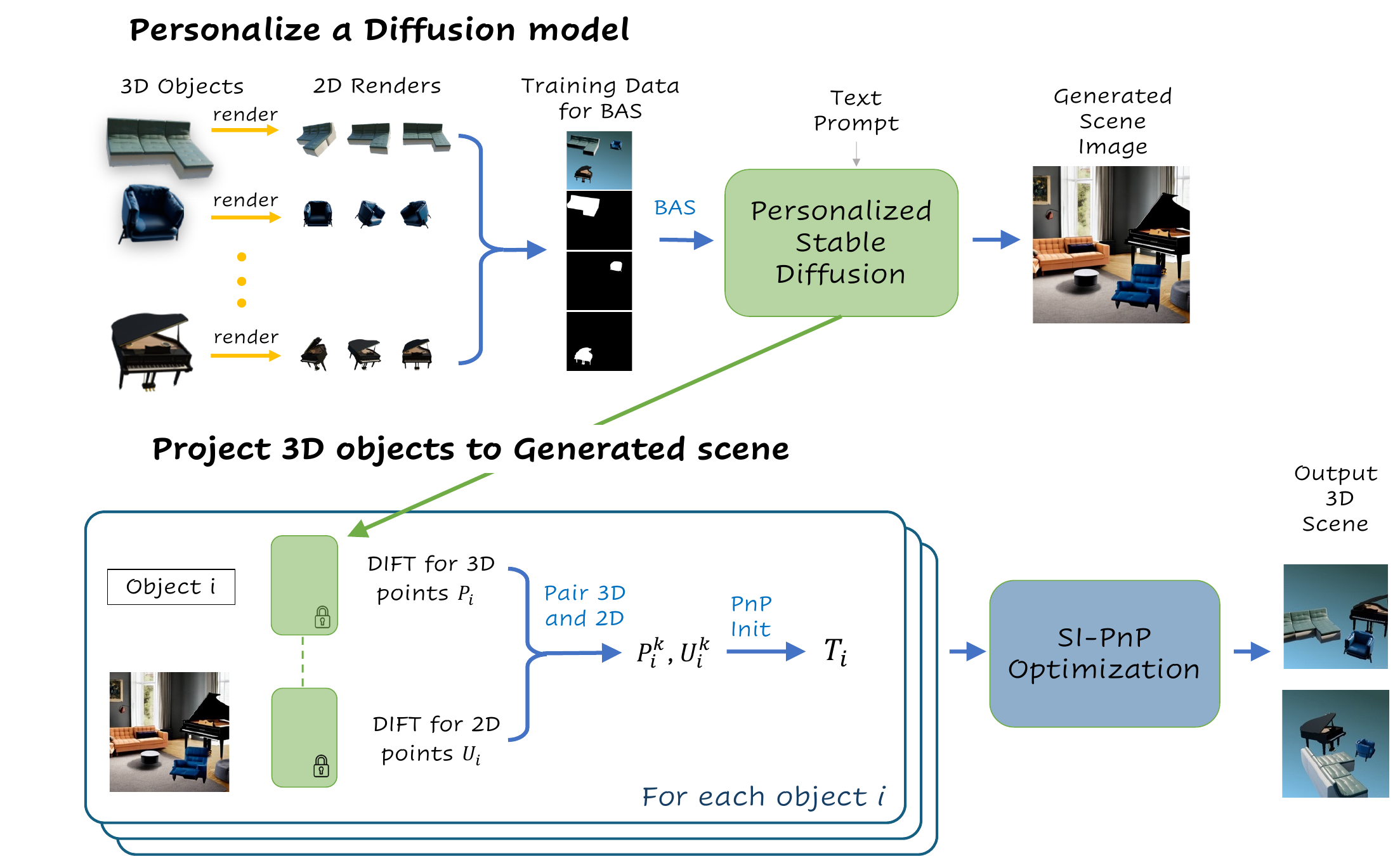}
    \caption{\ourmethod{} consists of two phases. First, given objects are used to personalize a text-to-image model, and a scene image is generated. In the second phase, we find a transformation $Tr_i$ for each 3D object $i$ to match the 2D arrangement presented in the generated scene image. $Tr_i$ is found using our \ourpnp{}, by matching the DIFT representation of objects and scene image.} %
    \label{fig:text23DarrangeArchitcture}
\end{figure}

Our method, \ourmethod{},  has two main stages: \textit{personalized image generation} and \textit{transformation optimization} (see Figure \ref{fig:text23DarrangeArchitcture} for an illustration).
In the first stage, we fine-tune a pretrained text-to-image model using rendered images of each object ${O_i}$ to generate a scene image $\cal{I}$ that includes all input objects, arranged to match a given description $y$.

In the second stage, we infer the 3D poses of the objects based on their appearances in $\cal{I}$. To determine these poses, we introduce a novel approach, \ourpnp{}, which finds a transformation $T_i$ that rotates and translates $O_i$ to match its appearance in $\cal{I}$.

The key idea behind \ourmethod{} is that text-to-image models inherently possess substantial information about the spatial arrangement of objects. This approach presents a dual challenge: generating an image of a scene with a specified set of objects and determining the 3D poses of these objects to match the generated scene image. The next section will describe these two steps in detail.

\subsection{Personalization by 3D objects} \label{sec:personalization}

Text-to-image personalization methods \cite{gal2022textual,ruiz2023dreambooth,avrahami2023break}, take a pre-trained diffusion model, like Stable Diffusion (SD), and a set of images of a concept to be personalized. They then fine-tune some parameters of the model so it can generate images of the personalized concept. Here, we personalize the model to generate a combined scene image containing all the given 3D objects organized according to the text-to-image prior knowledge, following a simple workflow. First, we render multiple images for each object from multiple viewpoints. We selected azimuths and elevations that are ``naturally found`` in the training set of SD, capturing mostly the front side of objects, and at a ``handheld`` elevation.  
Then, we created a single combined image, of multiple renderings, together with a set of masks indicating the location of each object in the combined image, see example Fig.\ref{fig:text23DarrangeArchitcture} in Training Data for BAS. We used Break-a-Scene (BAS) \cite{avrahami2023break} to finetune SD with these images.

Generating a scene image presented two main challenges: Handling a domain shift from natural images to rendered images, and addressing object neglect.
First, text-to-image models are trained on natural images, which often contain a natural background and meaningful context \cite{schuhmann2022laion}. In contrast, our rendered images have a monotonic background, and object arrangement should carry no meaning. This domain shift makes the personalization task more challenging. We used mask loss \cite{ryulow}, a masked version of the standard diffusion loss used in BAS, to alleviate the effect of this distribution shift, see more details in Appendix \ref{sec:MaskLoss}. This loss uses a combined mask of all objects to focus on the noise added only to the pixels of objects in the image. Second, diffusion models sometimes ignore nouns in their prompts, especially in long prompts \cite{attend-excite}. In our task, this can be detected quite easily since we match the input 3D objects with the generated images. This is discussed in \ref{subsec:object_neglect}.

\subsection{Optimizing Object Projection} \label{Object_Projection}
The second phase of our approach uses the image generated by the diffusion model to infer a plausible layout for our given objects. We achieve this by finding matching points between objects and their depictions in the generated scene image, and then use these matches to infer the relative pose of all objects.

\begin{figure}[t]
    \centering
    \includegraphics[width=0.8\textwidth]{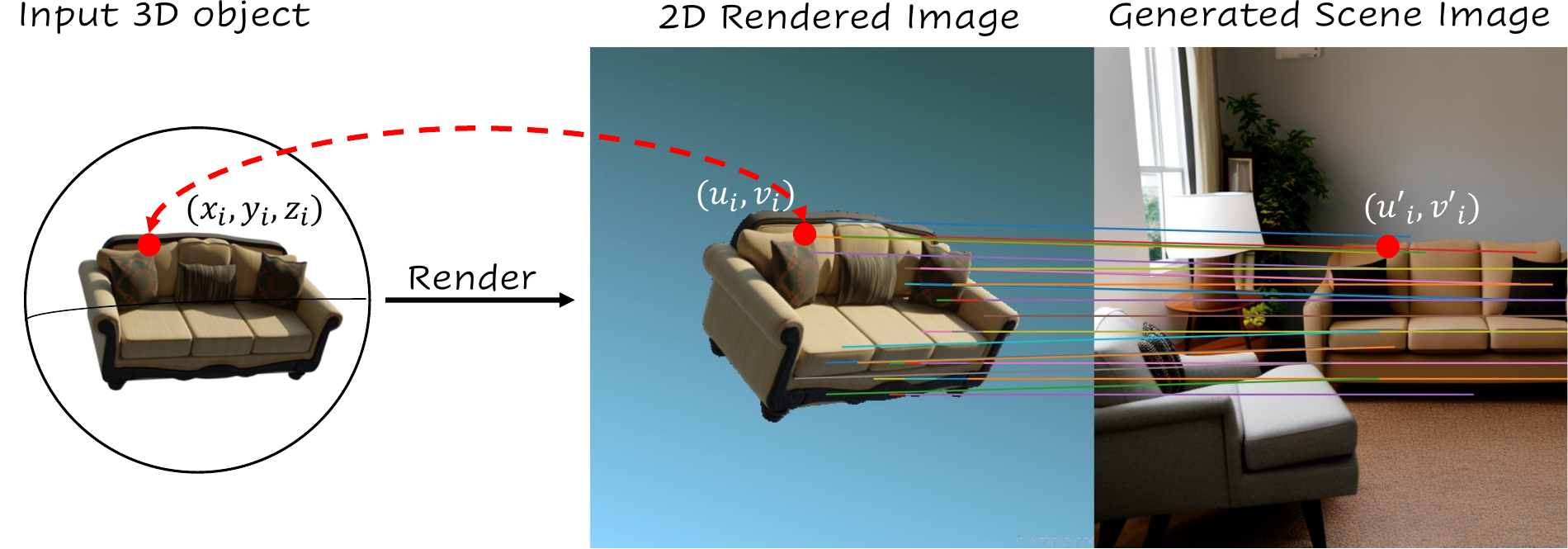}
    \caption{
    Establishing correspondences between keypoints of a 3D object and a scene image, by extracting the DIFT feature from the fine-tuned SD model. The process begins with rendering 2D images of a given object, used to map between features of the 3D object and the scene-generated image. On the right, each connecting line indicates a matched feature pair.
    }
    \label{fig:DIFTprocess}
\end{figure}

\textbf{Finding 3D-to-2D correspondence.} \label{subsec:find_correspondence} 
We wish to form a mapping between any given 3D object and its depiction in the generated scene image. Since we cannot establish correspondence between the 3D object and the scene image directly, we first render images of the objects from various viewpoints and use them to find correspondents using RANSAC \cite{fischler1981random} over DIFT \cite{DIFT}. RANSAC is a procedure based on iterative sampling that is robust to outliers. Once we identify the matches between the 2D renders and the scene image, we can project these matches back to the 3D object. Figure \ref{fig:DIFTprocess} illustrates this process. Our next goal is to determine the pose transformation of the 3D objects based on these keypoint matches.

\textbf{Perspective-n-Points.} \label{PnP} Finding a 3D pose of an object is a well-studied problem in computer graphics and is commonly solved using \textit{Perspective-n-Point} (PnP)~\cite{lu2018review}, a common method for estimating the position and orientation of a 3D object relative to a reference frame.
Formally, given a set of $M$ matches, PnP aims to find the rotation and translation of an object, represented by its 6 degrees of freedom, by analyzing 3D reference points of the $m^{th}$  match $P^m \in \mathbb{R}^{4}$ in homogeneous coordinates and their 2D image projections $U^m \in \mathbb{R}^{3}$, %
considering the camera's intrinsic parameters and perspective projection model:
\begin{equation}
    \label{problem-PnP}
    sU^m = K \ [R | T] \ P^m \quad.
\end{equation}
Here, $K$ is the matrix of intrinsic camera parameters, $s$ is a scale factor for the image point, $R$ and $T$ are the desired 3D rotation matrix and translation of the object, and $[R | T]$ is the composition of the rotation matrix and the translation vector to create the transformation matrix $Tr = [R | T]$.
Current methods for solving the PnP problem \eqref{problem-PnP} minimize the reprojection error to find the $R$ and $T$ matrices:
\begin{equation} \label{reprojection_error}
    \min_{R,T} \mathcal{L}_\mathrm{reprojection\ error} = \sum_{m=1,...,M} \lVert U^m - K \ [R | T] \ P^m / s \rVert_2^2 \quad.
\end{equation}
Here, $||\cdot||_2$ is the $L_2$ norm. This objective can be solved directly for $M=3$ matches, or by using RANSAC for $M>3$ matches to ensure robustness against outliers in the set of matches. Additionally, given an initialization, this objective can be efficiently optimized using standard solvers such as gradient descent.

\subsection{Side-Information PnP (\ourpnp)}
PnP described above can be optimized efficiently, but we find that the solutions in practice are often poor. Two issues hurt the quality of the inferred posses. First, the generated scene image may contain non-physical arrangements, where objects are slightly distorted or even intersecting. Second, there may be several different 3D poses and scalings that are consistent with a given 2D image. 

As an example, consider the "Pisa Tower" effect, where a 2D representation of a person seemingly holding the Pisa Tower does not reveal whether they are holding a miniature or standing far from the real tower. The original PnP method is designed for real-world images, where all objects have their true 3D object size. In our case, however, SD may generate objects in an image whose relative size is not consistent with the size of the 3D object. To address this mismatch, we developed \ourpnp{}, a novel method that incorporates physical world constraints to optimize object transformations in the scene.

To improve inferred posses we extend PnP to take into account prior information, by adding penalty terms to the loss of the PnP problem. We call our method \textit{\sidepnp{}} (\ourpnp).
Specifically, we added a term that encourages all objects to share a common "ground" surface, and a term that discourages objects from colliding with each other. Together, \ourpnp{} aims to balance three objectives: (1) The PnP objective. (2) Keep all objects on a shared surface. (3) Avoid collisions between objects. 
These penalties create a scene that not only matches the scene image but also yields well-structured object relations and sizes.

To take into account the case where all objects lay on a shared surface, we define a \textit{surface loss}. For each object $O_i$, we denote the 4 bottom vertices of its bounding box as $O_i^{\mathrm{floor}} \in \mathbb{R}^{4}$ in homogeneous coordinates. Moreover, we assume a floor plane with a homogeneous representation $L \in \mathbb{R}^{4}$. 
Ensuring all objects' bottom vertices lie on the same plane implies they rest on a common floor, initialized as the mean surface of these vertices. This requires that the bottom vertices of all objects satisfy the plane equation $P^TL=0$, where $P$ is a point in homogeneous representation \cite{hartley2003multiple}. The joint surface loss is defined as:
\begin{equation}
    \mathcal{L}_{\mathrm{surface}}= \sum_{i=1,...,N} \lVert ([R_i | T_i] \ O_i^{\mathrm{floor}})^T L \rVert_2^2.
\end{equation}

We also include a term for avoiding collisions between objects, by penalizing overlap between their 3D bounding boxes. The collision loss term is defined as:
\begin{equation}
    \mathcal{L}_{\mathrm{collision}} = \sum_{i=1,...,N} \sum_{j=1,...,N} \mathrm{IOU}([R_i | T_i] \ \mathrm{bbox}_i, [R_j | T_j] \ \mathrm{bbox}_j) \quad,
    \label{eq:collision}
\end{equation}
where $\mathrm{IOU}$ denotes the intersection-over-union between 3D bounding boxes, and $\mathrm{bbox}_i$ represents the 3D bounding box of the $i^{th}$ object.
Finally, we include the projection error from Eq.~\ref{reprojection_error} across all the $M$ 3D objects: 
\begin{equation*}
    \mathcal{L}_{\mathrm{reprojection}} = \sum_{i=1}^N \sum_{m=1}^M \lVert U_i^{m} - K \ [R_i | T_i] \ P_i^{m} / s \rVert_2^2\quad,
\end{equation*}
where, $\{(P_i^m, U_i^m)\}_{m=1,...,M}$ are the matching points and their corresponding image points described in Section~\ref{PnP}.
Thus, the total loss used in \ourpnp{} is:  
\begin{equation}
    \mathcal{L}_{\mathrm{SI-PNP}} = \mathcal{L}_{\mathrm{reprojection}} + \omega_{\mathrm{surface}}\mathcal{L}_{\mathrm{surface}} + \omega_{\mathrm{collision}}\mathcal{L}_{\mathrm{collision}} \quad,
    \label{eq:SI_PnP}
\end{equation}
Here, $\omega_{\mathrm{surface}}$ and  $\omega_{\mathrm{collision}}$ are hyperparameters that determine the relative weight of the losses.  $\mathcal{L}_{\mathrm{SI-PNP}}$ is optimized using standard gradient-based methods. 
Furthermore, we initialize \ourmethod{} using the transformations determined by the Perspective-n-Points (PnP) algorithm with the RANSAC procedure as described in \ref{PnP}, enhancing \ourpnp{} robustness to outliers.

\subsection{Handling Object neglect for More Objects in a scene}
One major limitation of current personalization methods is object neglect. This is the case where the text-to-image generative model fails to depict objects from its prompt in the generated image. It is a common problem in current diffusion models \cite{chefer2023attend}, and becomes worse as the number of objects in the prompt grows. We propose two methods to tackle this issue. First, we compute a "matching score" between the 3D objects and the generated scene, as described in \ref{sec:MatchingScore}. We use it to filter out scene images that do not contain all objects. This approach is quite effective for 2 - objects, but for more objects, the fraction of images to be filtered out becomes too large. 

Second, to address object neglect given a large number of objects, we propose an iterative approach.  We start by drawing two objects $O_1$ and $O_2$ uniformly at random from the set of objects $\{O_i \}_{i=1\ldots N}$, generate their scene image $\cal{I}_{\textnormal{12}}$, and determine their 3D relative position $\{T_i\}_{i=1,2}$. We then "merge" the two objects and treat their combined mesh as a single new object $O_{12}$. We then draw another object $O_3$ and repeat the process with two objects, the combined one $O_{12}$ and the new object $O_3$, to create a scene image $\cal{I}_{\textnormal{123}}$ and a new merged object $O_{123}$. We repeat the process until all objects are sampled and appear in the scene image.

We found that this iterative approach performs much better than personalizing a text-to-image model with multiple objects. This is because even when not all objects appear in the generated scene image $\cal{I}$, successful relations between some objects from previous iterations also indicate the positions of the new object in the scene. Combined with the ability of \ourpnp{} to prevent collision between objects, the iterative approach successfully yields a plausible arrangement of all objects.

\section{Experiments}

\subsection{Data and evaluation sets} \label{sets}
\begin{figure}[t] \label{Fig:2-objects}
    \centering
    \includegraphics[width=\textwidth]{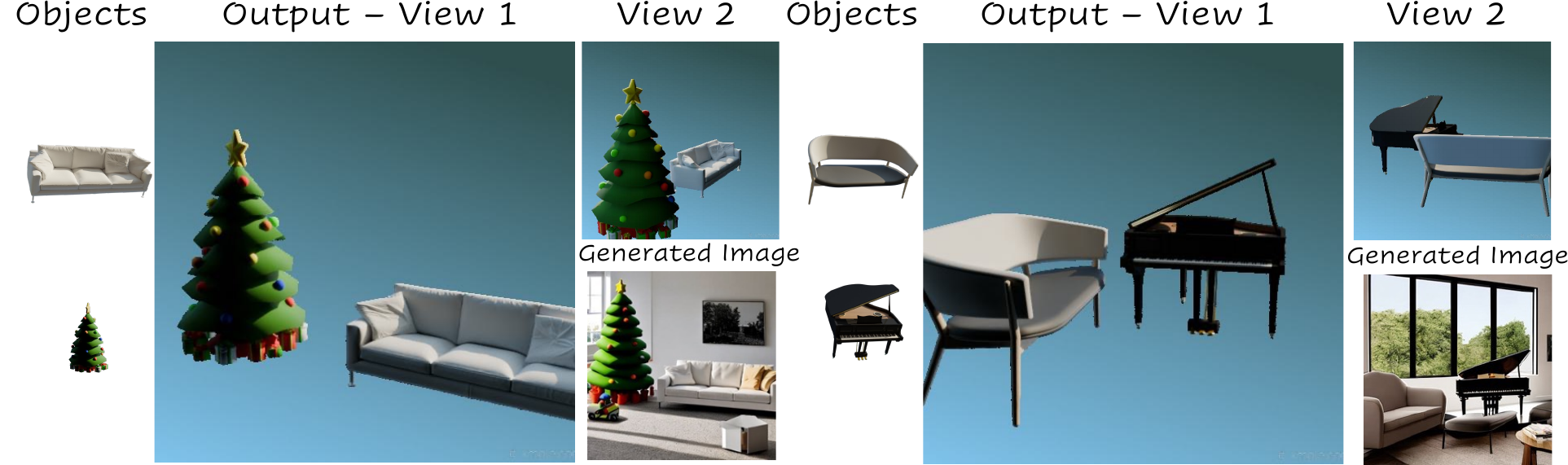}
    \caption{Qualitative examples of 2-object layouts generated by \ourmethod{}.} 
    \label{fig:3_objects}
\end{figure}

We created two evaluation sets, to quantify the quality of our approach, both composed of 3D furniture meshes taken from the Objaverse \cite{objaverse} dataset, a repository featuring over 800K high-quality 3D assets. 
(1) First, a \textbf{object-scene} set with 65 different scenes, each containing two objects, and of three scene categories: ``living room'', ``office'', ``dining table''. Shown in Fig. \ref{Fig:2-objects}.
(2) Second, a \textbf{multi-objects} to test the iterative approach. The set contains 70 different scenes, consisting of 3 to 5 objects, and of three scene categories: ``living room'', ``office'', ``dining table''.

\subsection{Baseline methods} \label{baselines}
We are not aware of previous work that addresses the open-set arrangement that this paper tackles. We therefore compare with two pre-defined heuristics:
\textbf{(1) Uniform. } We sample object positions uniformly at random as follows: Translation  $(x, y)$ in the range of $[-2, 2]$ and rotation parameters $\theta_z$ in the range of $[-180, 180]$ degrees. \textbf{(2) Circular. }  We place objects on a circle, all at equal distance from each other and facing the center of the circle. This heuristic arrangement is quite natural for furniture arrangement in rooms. 

Some scene generation methods expect a layout or a set of spatial relations among objects represented as scene graphs \cite{tang2024diffuscene,lin2024instructscene,Paschalidou2021NEURIPS}. These are not available in our setting. We were still interested in testing what layouts are generated by such methods, and if the spatial relations provided are lean. We used CommonScenes \cite{zhai2023commonscenes}, with an input scene graph that only had relations of type ``close by'' as an input. %

\textbf{Implementation details.}
When optimizing the transformation matrices, we used the Adam optimizer with a fixed learning rate of $0.001$ for 2000 steps.  For hyper parameters, we set $\omega_{\mathrm{surface}}=10^6$ and $\omega_{\mathrm{collision}}=10^4$ based on manual inspection in early development phases. See Appendix \ref{seq:implementation_details} for more details.

\section{Results}

\subsection{Layout quality - Automated methods} \label{seq: user study}

For each scene in the ``object-scenes'' set, we rendered a 2D image of the scene and used this image to calculate three repetitive scores: (1) Fréchet Inception Distance (FID) score \cite{heusel2017gans}, (2) Kernel Inception Distance (KID) Score\cite{binkowski2018demystifying}, and (3) Clip Similarity Score. The scores presented in Tables \ref{tab:fid_kid_comparison} and \ref{tab:clipscore_comparison} are the average scores of overall scenes from the same category. 
FID and KID scores were compared against the human-level arrangements found in the Objaverse dataset \cite{objaverse}. 
For more details, see Appendix \ref{Evaluation Metrics}.
As can be seen, \ourmethod{} achieved better average performance over all scene categories, and all three scores. For each score, our method also outperformed the baseline for the majority of the categories.

\begin{table}[ht]
    \centering
    \caption{\textbf{Realism of scene generation} evaluated through FID and KID scores between renders of generated and real scenes (lower FID and KID values preferred).}
    \scalebox{0.9}{
        \begin{tabular}{lcccccccc}
            \toprule
            \midrule
            \multicolumn{8}{c}{FID and KID Comparison Table} \\
            \midrule
            \textbf{Method}  & \multicolumn{2}{c}{\textbf{Livingroom}} & \multicolumn{2}{c}{\textbf{Office}} & \multicolumn{2}{c}{\textbf{Dining}} & \multicolumn{2}{c}{\textbf{Average}} \\
            \cline{2-9}
            & FID & KID & FID & KID & FID & KID & FID & KID \\
            \midrule
            CommonScenes & 286.5 & 5.4 & 305.7 & 10.2 & 324.1 & \textbf{16.5} & 305.43 & 10.7 \\
            \ourmethod{} (ours) & \textbf{282.3} & \textbf{5.2} & \textbf{301.7} & \textbf{9.0} & \textbf{302.8} & 16.7 & \textbf{295.60} & \textbf{10.3} \\
            \bottomrule
        \end{tabular}
    }
    \label{tab:fid_kid_comparison}
\end{table}

\begin{table}[ht]
    \centering
    \caption{\textbf{Fidelity of scene generation to scene descriptor} evaluated through ClipScore between renders of scenes generated by\ourmethod{} and CommonScenes (higher CLIP-score values preferred).}
    \scalebox{0.9}{
    \begin{tabular}{l c c c c}
         \toprule
         \midrule
        \multicolumn{5}{c}{CLIP-score Comparison Table} \\
        \midrule
        \textbf{Method}  & \textbf{Livingroom} & \textbf{Office} & \textbf{Dining} & \textbf{Average} \\
        \midrule
        CommonScenes & 23.42 & 25.57 & \textbf{24.81} & 24.60 \\
        \ourmethod{} (ours) & \textbf{25.53} & \textbf{27.07} & 23.95 & \textbf{25.52} \\
        \bottomrule
    \end{tabular}
    }
    \label{tab:clipscore_comparison}
\end{table}

\subsection{User study - comparison with Naive layouts }

We used raters from AMT to evaluate the quality of 3D layouts generated with \ourmethod{} compared to random and circular layouts. We used a 3-way alternative forced choice (3AFC) protocol. Raters were shown three scenes, one generated with our method and two with baselines, given two views of each scene. The preferred scene was determined by a majority vote of three raters. We calculated the standard error of the mean for each experiment. We evaluated both sets presented in \ref{sets}. For the first set, we used \ourmethod{}, while for the second set, we used the iterative \ourmethod{}. The results, shown in Table \ref{tab:user_study_random_circular}, indicate that \ourmethod{} outperformed the baselines for both sets.

We also compared \ourmethod{} with a graph-based method (CommonScenes \cite{zhai2023commonscenes}) using only "close-to" spatial relations as edges (see \ref{baselines}). Naturally, CommonScene would have been much better if meaningful spatial relations were available, but in our task, such relations are not provided.

\begin{table}[t]
    \centering
    \small
    \begin{tabular}{{c @{\hspace{30cm}} c}}
    ~\hspace{-15pt}~
        \begin{minipage}{0.54\textwidth}
            \centering
            \caption{\textbf{Layout quality user Study }. Percentage of scenes preferred by raters, in our two evaluation datasets. Error bars denote s.e.m.}
            \begin{tabular}{lccc}
                \toprule
                \midrule
                & \ourmethod{} & Uniform & Circular \\
                \midrule
                object-scene & \textbf{73.8} $\pm$ 2.4 & 18.4 $\pm$ 1.9 & 7.8 $\pm$ 0.8 \\
                multi-object & \textbf{50.0} $\pm$ 2.9 & 34.2 $\pm$ 2.7 & 15.8 $\pm$   1.6 \\
                \bottomrule
            \end{tabular}
            \label{tab:user_study_random_circular}
        \end{minipage} 
        \hfill
        ~\hspace{3pt}~
        \begin{minipage}{0.46\textwidth}
            \centering
            \small
            \caption{\textbf{Layout quality user Study}. As in Table 3. NaiveGraph is CommonScene with "close-to" edges. Error bars denote s.e.m.}
            \begin{tabular}{lcc}
                \toprule
                \midrule
                & \ourmethod{} & Naive Graph \\
                \midrule
                & \textbf{66.1} $\pm$ 2.7 & 33.9 $\pm$ 2.7 \\
                & \textbf{61.2} $\pm$ 2.8 & 38.7 $\pm$ 2.8 \\
                \bottomrule
            \end{tabular}
            \label{tab:user_study}
        \end{minipage}
    \end{tabular}
\end{table}

\begin{figure}[ht]
    \centering
    \includegraphics[width=1.\textwidth, trim={0.cm 0.cm 1.cm 0.cm},clip]{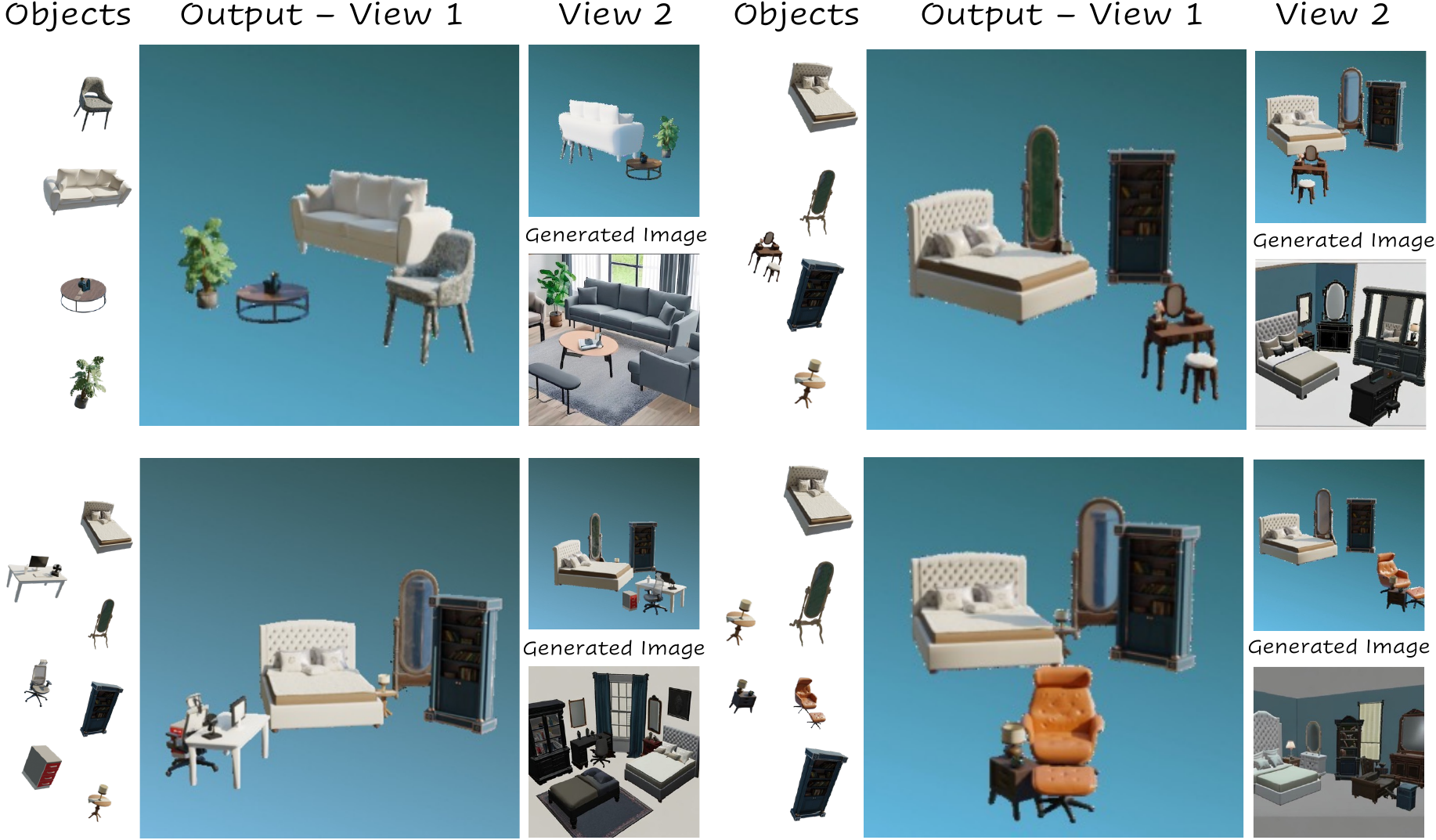}
    \caption{Iterative approach results: \ourmethod{} layout for multi-object scenes. We show scenes of a living room and a bedroom, each with multiple objects. We present scenes with 4 to 7 objects produced with the iterative approach.}
    \label{fig:ResultsIterativeApproach}
\end{figure}

\subsection{Qualitative Results}
Figure \ref{fig:3_objects} shows two scenes generated using \ourmethod{} for a pair of objects. Figure \ref{fig:ResultsIterativeApproach} demonstrates the effectiveness of our iterative approach in generating scenes with 4 to 7 objects.
More examples are given in the appendix.

\subsection{Ablation}
To assess the contribution of different components of \ourmethod{}, we evaluate our pipeline while removing some components. Table \ref{tab:user_study_personalization} looks into the personalization component. It 
shows that reaters strongly preferred layouts generated with the personalization component. Qualitative examples are given in Appendix \ref{fig:ablation_personalization}.
Table \ref{tab:user_study_SI-PNP} quantifies the contribution of \ourmethod{} compared with standard PnP. It shows that
the raters preferred the scenes with the \ourpnp{} optimization in $87.7\%$ of the scenes. 

\begin{table}[t]
    \centering
    \begin{minipage}{0.48\textwidth}
        \centering
        \caption{\textbf{Personalization ablation - User study.} The percentage of cases raters preferred a scene generated with and without personalization. Error bars denote s.e.m.}
        \begin{tabular}{lc}
             \toprule
             \midrule
              & percent preferred  \\
             \midrule
                w/ personalization & \textbf{81.5} $\pm$ 1.8 \\ 
                w/o personalization & 18.4 $\pm$ 1.8 \\
             \bottomrule
        \end{tabular}
        \label{tab:user_study_personalization}    
    \end{minipage}
    \hfill
    \begin{minipage}{0.48\textwidth}
    \centering
    \caption{\textbf{\ourpnp{} ablation - User study.} The percentage of cases raters preferred a scene generated with and without \ourpnp{}. Error bars denote s.e.m.}
    \begin{tabular}{lcc}
         \toprule
         \midrule
         & percent preferred  \\
          \midrule
          SI-PnP & \textbf{87.7} $\pm$ 1.3 \\
          PnP    & 12.3 $\pm$ 1.3 \\
         \bottomrule
    \end{tabular}
    \label{tab:user_study_SI-PNP}
    \end{minipage}
\end{table}

\begin{table}[h]
    \centering
\end{table}

\subsection{Object neglect}
\label{subsec:object_neglect}
\begin{figure}[h!]
    \centering
    \includegraphics[width=\textwidth]{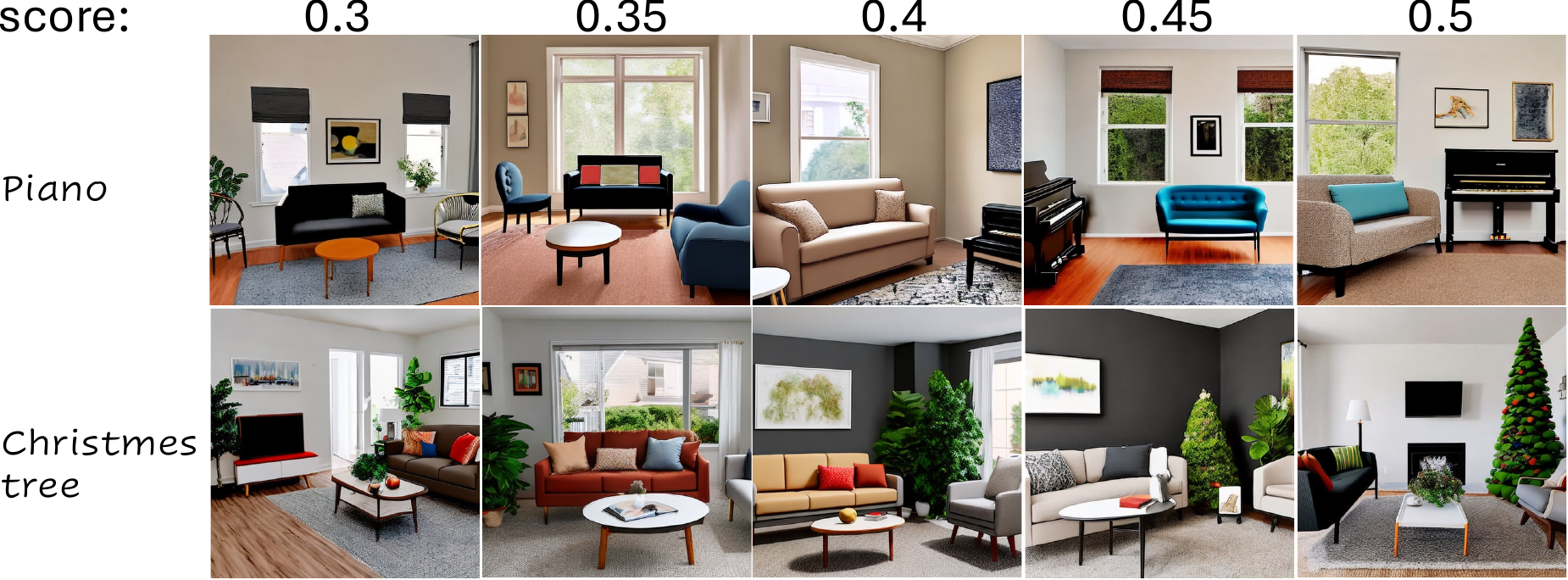}
    \caption{%
    Addressing object neglect through post-hoc filtering: Showing 5 matching scores (top) for two objects (right). Scores indicate object prominence in the generated image.
    } 
    \label{fig:dift-score}
\end{figure}

As a by-product of our matching process,  we compute a matching score that quantifies if a given object is present in an image. Figure \ref{fig:dift-score}, shows that smaller scores indicate the absence of the object of interest, and larger scores signify its presence. The higher the values, the more visually present the object is.

\section{Limitations}
\ourmethod{} has several limitations. Primarily, it depends on the underlying personalization process and inherits its limitations, such as object neglect, especially when a large number of objects need to be depicted. Personalization methods rarely succeed in generating more than four personalized objects. We address these limitations using post-filtering of scene images with neglected objects and our iterative method, and we expect that as personalization methods improve, \ourmethod{} will benefit. Additionally, we currently only model objects and not the surroundings. Finally, diffusion models may generate objects at incorrect scales, often confusing the pose estimation due to a "Pisa tower" effect.

\section{Conclusion}
We presented \ourmethod{}, an approach to arrange 3D objects into plausible layouts by leveraging text-to-image model personalization. \ourmethod{} offers a new way to distill the knowledge in text-to-image models. By personalizing the model with images of given 3D objects, we can generate images of plausible arrangements and then infer the 3D layout of objects using a modification of a PnP approach. This method allows for disentangling the generation of 3D objects from arranging them into scenes and offers new ways to use priors over the distribution of natural images for generating content.

\medskip

\newpage
{
\small
\bibliography{egbib}

\begin{thebibliography}{10}

\bibitem{avrahami2023break}
O.~Avrahami, K.~Aberman, O.~Fried, D.~Cohen{-}Or, and D.~Lischinski.
\newblock Break-{A}-{S}cene: {E}xtracting multiple concepts from a single
  image.
\newblock In {\em {SIGGRAPH} Asia 2023 Conference Papers, {SA} 2023, Sydney,
  NSW, Australia, December 12-15, 2023}, pages 96:1--96:12. {ACM}, 2023.

\bibitem{bahmani2023cc3d}
S.~Bahmani, J.~J. Park, D.~Paschalidou, X.~Yan, G.~Wetzstein, L.~Guibas, and
  A.~Tagliasacchi.
\newblock {CC3D}: {L}ayout-conditioned generation of compositional {3D} scenes.
\newblock {\em arXiv preprint arXiv:2303.12074}, 2023.

\bibitem{binkowski2018demystifying}
M.~Bi{\'n}kowski, D.~J. Sutherland, M.~Arbel, and A.~Gretton.
\newblock Demystifying mmd gans.
\newblock {\em arXiv preprint arXiv:1801.01401}, 2018.

\bibitem{chaudhuri2011probabilistic}
S.~Chaudhuri, E.~Kalogerakis, L.~Guibas, and V.~Koltun.
\newblock Probabilistic reasoning for assembly-based 3{D} modeling.
\newblock In {\em ACM SIGGRAPH 2011 papers}, pages 1--10, 2011.

\bibitem{chaudhuri2010data}
S.~Chaudhuri and V.~Koltun.
\newblock Data-driven suggestions for creativity support in 3{D} modeling.
\newblock In {\em ACM SIGGRAPH Asia 2010 papers}, pages 1--10, 2010.

\bibitem{chefer2023attendandexcite}
H.~Chefer, Y.~Alaluf, Y.~Vinker, L.~Wolf, and D.~Cohen-Or.
\newblock Attend-and-excite: Attention-based semantic guidance for
  text-to-image diffusion models, 2023.

\bibitem{attend-excite}
H.~Chefer, Y.~Alaluf, Y.~Vinker, L.~Wolf, and D.~Cohen-Or.
\newblock Attend-and-excite: Attention-based semantic guidance for
  text-to-image diffusion models.
\newblock {\em ACM Transactions on Graphics (TOG)}, 42(4):1--10, 2023.

\bibitem{chefer2023attend}
H.~Chefer, Y.~Alaluf, Y.~Vinker, L.~Wolf, and D.~Cohen-Or.
\newblock Attend-and-excite: Attention-based semantic guidance for
  text-to-image diffusion models.
\newblock {\em ACM Transactions on Graphics (TOG)}, 42(4):1--10, 2023.

\bibitem{chen2022tango}
Y.~Chen, R.~Chen, J.~Lei, Y.~Zhang, and K.~Jia.
\newblock {TANGO}: Text-driven photorealistic and robust 3{D} stylization via
  lighting decomposition.
\newblock {\em Advances in Neural Information Processing Systems},
  35:30923--30936, 2022.

\bibitem{objaverse}
M.~Deitke, D.~Schwenk, J.~Salvador, L.~Weihs, O.~Michel, E.~VanderBilt,
  L.~Schmidt, K.~Ehsani, A.~Kembhavi, and A.~Farhadi.
\newblock Objaverse: A universe of annotated 3d objects.
\newblock In {\em Proceedings of the IEEE/CVF Conference on Computer Vision and
  Pattern Recognition}, pages 13142--13153, 2023.

\bibitem{devaranjan2020meta}
J.~Devaranjan, A.~Kar, and S.~Fidler.
\newblock Meta-{S}im2: {U}nsupervised learning of scene structure for synthetic
  data generation.
\newblock In {\em Computer Vision--ECCV 2020: 16th European Conference,
  Glasgow, UK, August 23--28, 2020, Proceedings, Part XVII 16}, pages 715--733.
  Springer, 2020.

\bibitem{dhamo2021graph}
H.~Dhamo, F.~Manhardt, N.~Navab, and F.~Tombari.
\newblock Graph-to-3d: End-to-end generation and manipulation of 3d scenes
  using scene graphs.
\newblock In {\em Proceedings of the IEEE/CVF International Conference on
  Computer Vision}, pages 16352--16361, 2021.

\bibitem{fischler1981random}
M.~A. Fischler and R.~C. Bolles.
\newblock Random sample consensus: a paradigm for model fitting with
  applications to image analysis and automated cartography.
\newblock {\em Communications of the ACM}, 24(6):381--395, 1981.

\bibitem{fisher2012example}
M.~Fisher, D.~Ritchie, M.~Savva, T.~Funkhouser, and P.~Hanrahan.
\newblock Example-based synthesis of 3d object arrangements.
\newblock {\em ACM Transactions on Graphics (TOG)}, 31(6):1--11, 2012.

\bibitem{funkhouser2004modeling}
T.~Funkhouser, M.~Kazhdan, P.~Shilane, P.~Min, W.~Kiefer, A.~Tal,
  S.~Rusinkiewicz, and D.~Dobkin.
\newblock Modeling by example.
\newblock {\em ACM transactions on graphics (TOG)}, 23(3):652--663, 2004.

\bibitem{make-a-scene}
O.~Gafni, A.~Polyak, O.~Ashual, S.~Sheynin, D.~Parikh, and Y.~Taigman.
\newblock Make-a-scene: Scene-based text-to-image generation with human priors.
\newblock In {\em European Conference on Computer Vision}, pages 89--106.
  Springer, 2022.

\bibitem{gafni2022make}
O.~Gafni, A.~Polyak, O.~Ashual, S.~Sheynin, D.~Parikh, and Y.~Taigman.
\newblock Make-{A}-{S}cene: {S}cene-based text-to-image generation with human
  priors.
\newblock In {\em European Conference on Computer Vision}, pages 89--106.
  Springer, 2022.

\bibitem{TextualInversion}
R.~Gal, Y.~Alaluf, Y.~Atzmon, O.~Patashnik, A.~H. Bermano, G.~Chechik, and
  D.~Cohen-or.
\newblock An image is worth one word: Personalizing text-to-image generation
  using textual inversion.
\newblock In {\em The Eleventh International Conference on Learning
  Representations}, 2022.

\bibitem{gal2022textual}
R.~Gal, Y.~Alaluf, Y.~Atzmon, O.~Patashnik, A.~H. Bermano, G.~Chechik, and
  D.~Cohen-Or.
\newblock An image is worth one word: Personalizing text-to-image generation
  using textual inversion, 2022.

\bibitem{gan2023idesigner}
R.~Gan, X.~Wu, J.~Lu, Y.~Tian, D.~Zhang, Z.~Wu, R.~Sun, C.~Liu, J.~Zhang,
  P.~Zhang, et~al.
\newblock idesigner: A high-resolution and complex-prompt following
  text-to-image diffusion model for interior design.
\newblock {\em arXiv preprint arXiv:2312.04326}, 2023.

\bibitem{hartley2003multiple}
R.~Hartley and A.~Zisserman.
\newblock {\em Multiple view geometry in computer vision}.
\newblock Cambridge university press, 2003.

\bibitem{heusel2017gans}
M.~Heusel, H.~Ramsauer, T.~Unterthiner, B.~Nessler, and S.~Hochreiter.
\newblock Gans trained by a two time-scale update rule converge to a local nash
  equilibrium.
\newblock {\em Advances in neural information processing systems}, 30, 2017.

\bibitem{hollein2023text2room}
L.~H{\"o}llein, A.~Cao, A.~Owens, J.~Johnson, and M.~Nie{\ss}ner.
\newblock Text2room: Extracting textured 3d meshes from 2d text-to-image
  models.
\newblock {\em arXiv preprint arXiv:2303.11989}, 2023.

\bibitem{huang2015single}
Q.~Huang, H.~Wang, and V.~Koltun.
\newblock Single-view reconstruction via joint analysis of image and shape
  collections.
\newblock {\em ACM Trans. Graph.}, 34(4):87--1, 2015.

\bibitem{jain2022zeroshot}
A.~Jain, B.~Mildenhall, J.~T. Barron, P.~Abbeel, and B.~Poole.
\newblock Zero-shot text-guided object generation with dream fields, 2022.

\bibitem{jiang2012learning}
Y.~Jiang, M.~Lim, and A.~Saxena.
\newblock Learning object arrangements in {3D} scenes using human context.
\newblock In {\em Proceedings of the 29th International Coference on
  International Conference on Machine Learning}, pages 907--914, 2012.

\bibitem{jiang20233d}
Z.~Jiang, G.~Lu, X.~Liang, J.~Zhu, W.~Zhang, X.~Chang, and H.~Xu.
\newblock {3D-TOGO}: Towards text-guided cross-category 3{D} object generation.
\newblock In {\em Proceedings of the AAAI Conference on Artificial
  Intelligence}, volume~37, pages 1051--1059, 2023.

\bibitem{jyothi2019layoutvae}
A.~A. Jyothi, T.~Durand, J.~He, L.~Sigal, and G.~Mori.
\newblock Layout{VAE}: {S}tochastic scene layout generation from a label set.
\newblock In {\em Proceedings of the IEEE/CVF International Conference on
  Computer Vision}, pages 9895--9904, 2019.

\bibitem{kalogerakis2012probabilistic}
E.~Kalogerakis, S.~Chaudhuri, D.~Koller, and V.~Koltun.
\newblock A probabilistic model for component-based shape synthesis.
\newblock {\em Acm Transactions on Graphics (TOG)}, 31(4):1--11, 2012.

\bibitem{kreavoy2007model}
V.~Kreavoy, D.~Julius, and A.~Sheffer.
\newblock Model composition from interchangeable components.
\newblock In {\em 15th Pacific Conference on Computer Graphics and Applications
  (PG'07)}, pages 129--138. IEEE, 2007.

\bibitem{lee2022understanding}
H.-H. Lee and A.~X. Chang.
\newblock Understanding pure {CLIP} guidance for voxel grid {N}e{RF} models.
\newblock {\em arXiv preprint arXiv:2209.15172}, 2022.

\bibitem{lin2024instructscene}
C.~Lin and Y.~Mu.
\newblock Instructscene: Instruction-driven 3d indoor scene synthesis with
  semantic graph prior.
\newblock In {\em International Conference on Learning Representations (ICLR)},
  2024.

\bibitem{lin2023magic3d}
C.-H. Lin, J.~Gao, L.~Tang, T.~Takikawa, X.~Zeng, X.~Huang, K.~Kreis,
  S.~Fidler, M.-Y. Liu, and T.-Y. Lin.
\newblock Magic3d: High-resolution text-to-3d content creation.
\newblock In {\em Proceedings of the IEEE/CVF Conference on Computer Vision and
  Pattern Recognition}, pages 300--309, 2023.

\bibitem{liu2014creating}
T.~Liu, S.~Chaudhuri, V.~G. Kim, Q.~Huang, N.~J. Mitra, and T.~Funkhouser.
\newblock Creating consistent scene graphs using a probabilistic grammar.
\newblock {\em ACM Transactions on Graphics (TOG)}, 33(6):1--12, 2014.

\bibitem{lu2018review}
X.~X. Lu.
\newblock A review of solutions for perspective-n-point problem in camera pose
  estimation.
\newblock In {\em Journal of Physics: Conference Series}, volume 1087, page
  052009. IOP Publishing, 2018.

\bibitem{melas2023realfusion}
L.~Melas-Kyriazi, I.~Laina, C.~Rupprecht, and A.~Vedaldi.
\newblock Real{F}usion: 360° reconstruction of any object from a single image.
\newblock In {\em Proceedings of the IEEE/CVF Conference on Computer Vision and
  Pattern Recognition}, pages 8446--8455, 2023.

\bibitem{merrell2010computer}
P.~Merrell, E.~Schkufza, and V.~Koltun.
\newblock Computer-generated residential building layouts.
\newblock In {\em ACM SIGGRAPH Asia 2010 papers}, pages 1--12, 2010.

\bibitem{metzer2023latent}
G.~Metzer, E.~Richardson, O.~Patashnik, R.~Giryes, and D.~Cohen-Or.
\newblock Latent-{N}e{RF} for shape-guided generation of 3{D} shapes and
  textures.
\newblock In {\em Proceedings of the IEEE/CVF Conference on Computer Vision and
  Pattern Recognition}, pages 12663--12673, 2023.

\bibitem{michel2022text2mesh}
O.~Michel, R.~Bar-On, R.~Liu, S.~Benaim, and R.~Hanocka.
\newblock Text2{MESH}: Text-driven neural stylization for meshes.
\newblock In {\em Proceedings of the IEEE/CVF Conference on Computer Vision and
  Pattern Recognition}, pages 13492--13502, 2022.

\bibitem{mohammad2022clip}
N.~Mohammad~Khalid, T.~Xie, E.~Belilovsky, and T.~Popa.
\newblock {CLIP-MESH}: Generating textured meshes from text using pretrained
  image-text models.
\newblock In {\em SIGGRAPH Asia 2022 conference papers}, pages 1--8, 2022.

\bibitem{park2019semantic}
T.~Park, M.-Y. Liu, T.-C. Wang, and J.-Y. Zhu.
\newblock Semantic image synthesis with spatially-adaptive normalization.
\newblock In {\em Proceedings of the IEEE/CVF conference on computer vision and
  pattern recognition}, pages 2337--2346, 2019.

\bibitem{Paschalidou2021NEURIPS}
D.~Paschalidou, A.~Kar, M.~Shugrina, K.~Kreis, A.~Geiger, and S.~Fidler.
\newblock Atiss: Autoregressive transformers for indoor scene synthesis.
\newblock In {\em Advances in Neural Information Processing Systems (NeurIPS)},
  2021.

\bibitem{poole2022dreamfusion}
B.~Poole, A.~Jain, J.~T. Barron, and B.~Mildenhall.
\newblock Dreamfusion: Text-to-3d using 2d diffusion.
\newblock In {\em The Eleventh International Conference on Learning
  Representations}, 2022.

\bibitem{purkait2020sg}
P.~Purkait, C.~Zach, and I.~Reid.
\newblock {SG-VAE}: {S}cene grammar variational autoencoder to generate new
  indoor scenes.
\newblock In {\em European Conference on Computer Vision}, pages 155--171.
  Springer, 2020.

\bibitem{radford2021learning}
A.~Radford, J.~W. Kim, C.~Hallacy, A.~Ramesh, G.~Goh, S.~Agarwal, G.~Sastry,
  A.~Askell, P.~Mishkin, J.~Clark, et~al.
\newblock Learning transferable visual models from natural language
  supervision.
\newblock In {\em International conference on machine learning}, pages
  8748--8763. PMLR, 2021.

\bibitem{raj2023dreambooth3d}
A.~Raj, S.~Kaza, B.~Poole, M.~Niemeyer, N.~Ruiz, B.~Mildenhall, S.~Zada,
  K.~Aberman, M.~Rubinstein, J.~Barron, Y.~Li, and V.~Jampani.
\newblock Dreambooth3d: Subject-driven text-to-3d generation, 2023.

\bibitem{rassin2023linguistic}
R.~Rassin, E.~Hirsch, D.~Glickman, S.~Ravfogel, Y.~Goldberg, and G.~Chechik.
\newblock Linguistic binding in diffusion models: Enhancing attribute
  correspondence through attention map alignment, 2023.

\bibitem{richardson2023texture}
E.~Richardson, G.~Metzer, Y.~Alaluf, R.~Giryes, and D.~Cohen-Or.
\newblock {TEXT}ure: Text-guided texturing of 3{D} shapes.
\newblock {\em arXiv preprint arXiv:2302.01721}, 2023.

\bibitem{rombach2022high}
R.~Rombach, A.~Blattmann, D.~Lorenz, P.~Esser, and B.~Ommer.
\newblock High-resolution image synthesis with latent diffusion models.
\newblock In {\em Proceedings of the IEEE/CVF conference on computer vision and
  pattern recognition}, pages 10684--10695, 2022.

\bibitem{StableDiffusion}
R.~Rombach, A.~Blattmann, D.~Lorenz, P.~Esser, and B.~Ommer.
\newblock High-resolution image synthesis with latent diffusion models.
\newblock In {\em Proceedings of the IEEE/CVF conference on computer vision and
  pattern recognition}, pages 10684--10695, 2022.

\bibitem{ruiz2023dreambooth}
N.~Ruiz, Y.~Li, V.~Jampani, Y.~Pritch, M.~Rubinstein, and K.~Aberman.
\newblock Dream{B}ooth: Fine tuning text-to-image diffusion models for
  subject-driven generation.
\newblock In {\em Proceedings of the IEEE/CVF Conference on Computer Vision and
  Pattern Recognition}, pages 22500--22510, 2023.

\bibitem{ryulow}
S.~Ryu.
\newblock Low-rank adaptation for fast text-to-image diffusion fine-tuning.
\newblock \url{https://github.com/cloneofsimo/lora}, 2023.

\bibitem{saharia2022photorealistic}
C.~Saharia, W.~Chan, S.~Saxena, L.~Li, J.~Whang, E.~L. Denton, K.~Ghasemipour,
  R.~Gontijo~Lopes, B.~Karagol~Ayan, T.~Salimans, et~al.
\newblock Photorealistic text-to-image diffusion models with deep language
  understanding.
\newblock {\em Advances in Neural Information Processing Systems},
  35:36479--36494, 2022.

\bibitem{schuhmann2022laion}
C.~Schuhmann, R.~Beaumont, R.~Vencu, C.~Gordon, R.~Wightman, M.~Cherti,
  T.~Coombes, A.~Katta, C.~Mullis, M.~Wortsman, et~al.
\newblock Laion-5b: An open large-scale dataset for training next generation
  image-text models.
\newblock {\em Advances in Neural Information Processing Systems},
  35:25278--25294, 2022.

\bibitem{shen2012structure}
C.-H. Shen, H.~Fu, K.~Chen, and S.-M. Hu.
\newblock Structure recovery by part assembly.
\newblock {\em ACM Transactions on Graphics (TOG)}, 31(6):1--11, 2012.

\bibitem{SongCXKTYY23}
L.~Song, L.~Cao, H.~Xu, K.~Kang, F.~Tang, J.~Yuan, and Y.~Zhao.
\newblock Roomdreamer: Text-driven 3d indoor scene synthesis with coherent
  geometry and texture.
\newblock {\em arXiv preprint arXiv:2305.11337}, 2023.

\bibitem{tang2023diffuscene}
J.~Tang, Y.~Nie, L.~Markhasin, A.~Dai, J.~Thies, and M.~Nie{\ss}ner.
\newblock Diffuscene: Scene graph denoising diffusion probabilistic model for
  generative indoor scene synthesis.
\newblock {\em arXiv preprint arXiv:2303.14207}, 2023.

\bibitem{tang2024diffuscene}
J.~Tang, Y.~Nie, L.~Markhasin, A.~Dai, J.~Thies, and M.~Nießner.
\newblock Diffuscene: Denoising diffusion models for generative indoor scene
  synthesis, 2024.

\bibitem{DIFT}
L.~Tang, M.~Jia, Q.~Wang, C.~P. Phoo, and B.~Hariharan.
\newblock Emergent correspondence from image diffusion.
\newblock In {\em Thirty-seventh Conference on Neural Information Processing
  Systems}, 2023.

\bibitem{tulsiani2018factoring}
S.~Tulsiani, S.~Gupta, D.~F. Fouhey, A.~A. Efros, and J.~Malik.
\newblock Factoring shape, pose, and layout from the 2{D} image of a 3{D}
  scene.
\newblock In {\em Proceedings of the IEEE Conference on Computer Vision and
  Pattern Recognition}, pages 302--310, 2018.

\bibitem{wang2023nerf}
C.~Wang, R.~Jiang, M.~Chai, M.~He, D.~Chen, and J.~Liao.
\newblock Ne{RF}-{A}rt: Text-driven neural radiance fields stylization.
\newblock {\em IEEE Transactions on Visualization and Computer Graphics}, 2023.

\bibitem{wang2023score}
H.~Wang, X.~Du, J.~Li, R.~A. Yeh, and G.~Shakhnarovich.
\newblock Score {J}acobian chaining: Lifting pretrained 2{D} diffusion models
  for 3{D} generation.
\newblock In {\em Proceedings of the IEEE/CVF Conference on Computer Vision and
  Pattern Recognition}, pages 12619--12629, 2023.

\bibitem{wang2018deep}
K.~Wang, M.~Savva, A.~X. Chang, and D.~Ritchie.
\newblock Deep convolutional priors for indoor scene synthesis.
\newblock {\em ACM Transactions on Graphics (TOG)}, 37(4):1--14, 2018.

\bibitem{wei2023lego}
Q.~A. Wei, S.~Ding, J.~J. Park, R.~Sajnani, A.~Poulenard, S.~Sridhar, and
  L.~Guibas.
\newblock Lego-net: Learning regular rearrangements of objects in rooms.
\newblock In {\em Proceedings of the IEEE/CVF Conference on Computer Vision and
  Pattern Recognition}, pages 19037--19047, 2023.

\bibitem{xu2012fit}
K.~Xu, H.~Zhang, D.~Cohen-Or, and B.~Chen.
\newblock Fit and diverse: {S}et evolution for inspiring 3{D} shape galleries.
\newblock {\em ACM Transactions on Graphics (TOG)}, 31(4):1--10, 2012.

\bibitem{yang2021indoor}
M.-J. Yang, Y.-X. Guo, B.~Zhou, and X.~Tong.
\newblock Indoor scene generation from a collection of semantic-segmented depth
  images.
\newblock In {\em Proceedings of the IEEE/CVF International Conference on
  Computer Vision}, pages 15203--15212, 2021.

\bibitem{yu2011make}
L.~F. Yu, S.~K. Yeung, C.~K. Tang, D.~Terzopoulos, T.~F. Chan, and S.~J. Osher.
\newblock Make it home: automatic optimization of furniture arrangement.
\newblock {\em ACM Transactions on Graphics (TOG)-Proceedings of ACM SIGGRAPH
  2011, v. 30,(4), July 2011, article no. 86}, 30(4), 2011.

\bibitem{zhai2023commonscenes}
G.~Zhai, E.~P. {\"O}rnek, S.-C. Wu, Y.~Di, F.~Tombari, N.~Navab, and B.~Busam.
\newblock Commonscenes: Generating commonsense 3d indoor scenes with scene
  graphs.
\newblock {\em arXiv preprint arXiv:2305.16283}, 2023.

\bibitem{quaternions}
F.~Zhang.
\newblock Quaternions and matrices of quaternions.
\newblock {\em Linear algebra and its applications}, 251:21--57, 1997.

\bibitem{zhao2021luminous}
Y.~Zhao, K.~Lin, Z.~Jia, Q.~Gao, G.~Thattai, J.~Thomason, and G.~S. Sukhatme.
\newblock Luminous: Indoor scene generation for embodied ai challenges.
\newblock {\em arXiv preprint arXiv:2111.05527}, 2021.

\end{thebibliography}
\bibliographystyle{abbrv}
}

\newpage
\appendix

\section{Appendix / supplemental material}

\subsection{Rendering Objects in the Scene}
We used Blender\footnote{https://www.blender.org/} with HDRI environment lighting\footnote{https://hdri-haven.com} to render 3D objects for BAS input data. Objects in the dataset might not match real-world scales. This scale mismatch can confuse the generative model. Therefore, we render multiple images for each object and apply a pre-processing step. In this step, we resize objects to fill their image areas while maintaining proportionality with other objects and real-world sizes.

\subsection{Personalization}
Despite optimizing object tokens and model weights with mask loss (see Section 3.2), background pixels may still appear in object edge patches. To address the domain shift from natural to rendered images, we add a background image to the combined images to enhance the natural appearance. We also use the prior preservation loss from Dreambooth \cite{ruiz2023dreambooth}. This loss reinforces the model's understanding of object classes using additional class-generated images.

To further counter the object neglect problem, and the tendency of our model to merge concepts, especially in visually similar objects, we use BAS's cross-attention loss\cite{avrahami2023break}. This loss measures the mean squared error deviation of the new token's cross-attention map from the concept mask. We apply this loss in both training phases to focus each new concept token on its concept's image region. However, we found that the most effective way to handle object neglect is calculating the Matching Score (Section~\ref{sec:MatchingScore}) and using it to indicate object presence in the scene image.

\subsection{Matching Score}\label{sec:MatchingScore}

\textbf{Finding Correspondences:} We extract feature descriptors from multiple 2D-rendered images of each object and the scene-generated image. We use the DIFT method for feature extraction \cite{DIFT}. Specifically, we input a noisy version of the image into the fine-tuned SD model from the personalization phase. We then extract intermediate layer activations as feature descriptors. The diffusion process is conditioned by the learned text tokens, such as "a photo of $<$asset$_i$$>$". We remove feature descriptors corresponding to the background of the rendered image.
For each feature vector in the rendered images, we use cosine distance to identify the closest matching keypoint in the scene-generated image. This establishes a correspondence pair $(u, v), (u', v')$, linking each point in the rendered image to its counterpart in the target image. We map each keypoint in the 2D rendered image to its 3D $(x,y,z)$ coordinate, based on the known projection used during rendering.
For each pair of corresponding keypoints, we record the cosine similarity score. In the next phase, we apply PnP with RANSAC to find the transformation for each object, where it tries to find the largest group of points that consensus about the same transformation, which are the "inlier keypoints". The median cosine similarity score of these keypoints is the "Matching score". We use this score to filter out scene images with the object neglect problem.

\subsection{Mask loss}\label{sec:MaskLoss}
The mask loss computes the loss of pixels indicated by the mask only. Therefore, the irrelevant background of the render does not affect the optimization process.

\begin{equation}
    \mathcal{L}_{\text{mask loss}} = \mathbb{E}_{z,s,\epsilon \sim \mathcal{N}(0,1),t} \left\| \epsilon \odot M - \epsilon_{\theta}(z_t, t, y) \odot M \right\|_2^2
\end{equation}

Here, $z_t$ represents the noisy latent at time step $t$, $y$ denotes the text prompt, $M$ is the mask, $\epsilon$ stands for the added noise, and $\epsilon \theta$ refers to the denoising network.

\subsection{implementation details} \label{seq:implementation_details}

We used BAS for personalizing stable-diffusion-2-1-base. Initialed-text-tokens were set to be the class of the object, for example, "chair" or "sofa". We used the prompt of this format  "A photo of a <scene description> with a <asset0> next to <asset1>". Prior preservation weight was set to 1. We also changed the BAS dataloader such that it will get multiple images from different viewpoints instead of a single image. The rest of the hyperparameters were identical to the default values in BAS.

\textbf{\ourpnp{}}
We optimize the translations and rotations separately, where the translations have been directly optimized, while the rotations have been converted into quaternions \cite{quaternions} which were optimized. The conversion to quaternions is necessary to keep the rotation as a unitary matrix to avoid object deformations.
In each step, we assembled the transform matrix as ${Tr_i = [R_i | T_i]}$ and used it to calculate the losses.

\subsection{Experiments Compute Resources}
We evaluated our method using an A100 GPU, each iteration took 20 minutes to produce a plausible scene. This time can be reduced using less generated examples.

\subsection{Experiments instructions}

We provided the raters with the following screen, and we asked them to choose the better scene. Each task took about 20 seconds, with a payment of 10 \textcent, equating to $18$/hour.

\begin{figure}[ht]
    \centering
    \includegraphics[width=\textwidth]{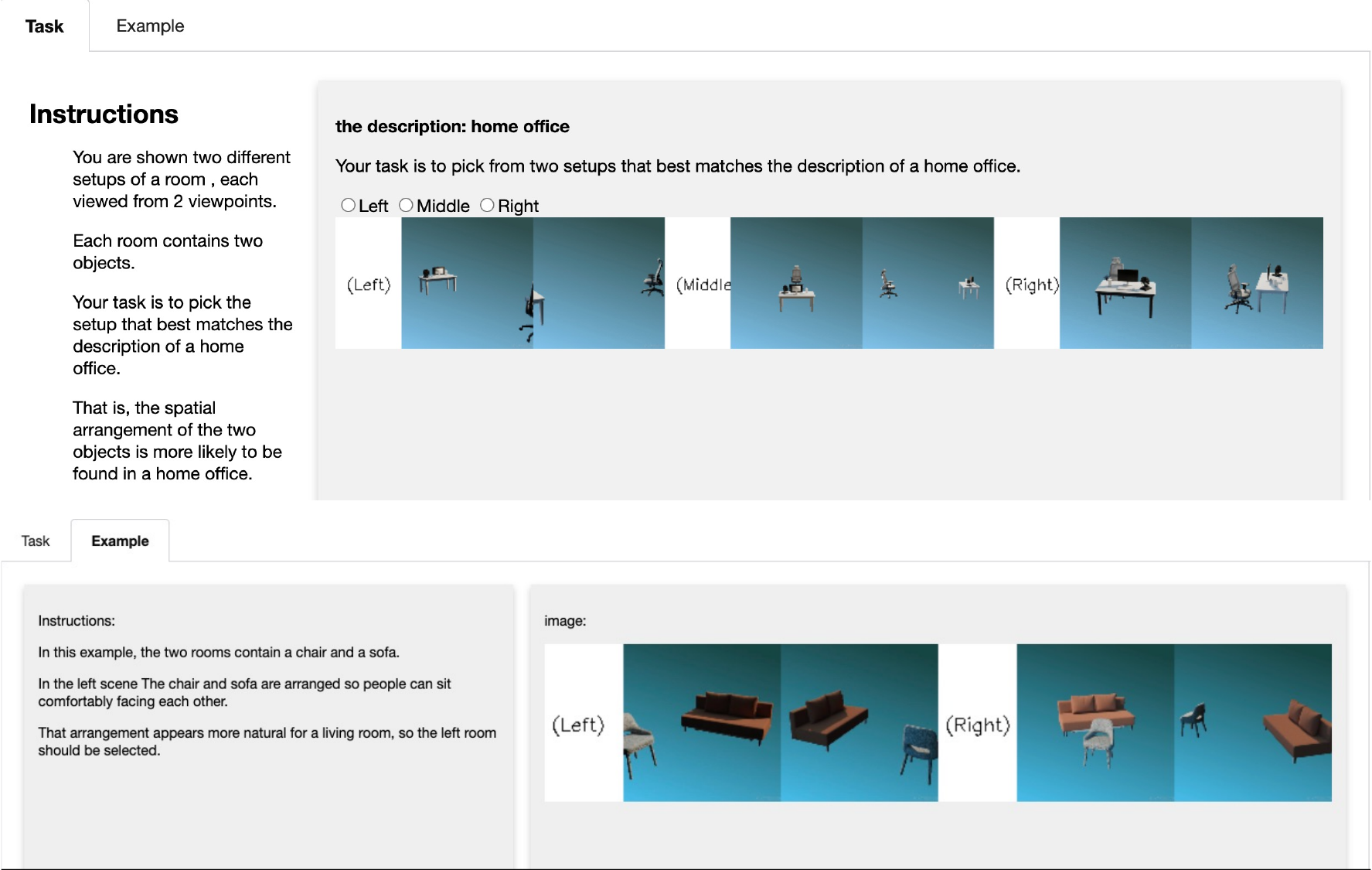}
    \caption{the screen on the top is the selection one, while the bottom presents the example. in each task, the raters were given 3 layouts, and were asked to choose the better layout}
\end{figure}

\section{Ablation}

\subsection{Personalization} \label{appendix-personalization}

\begin{figure}[ht]
    \centering
    \includegraphics[width=\textwidth]{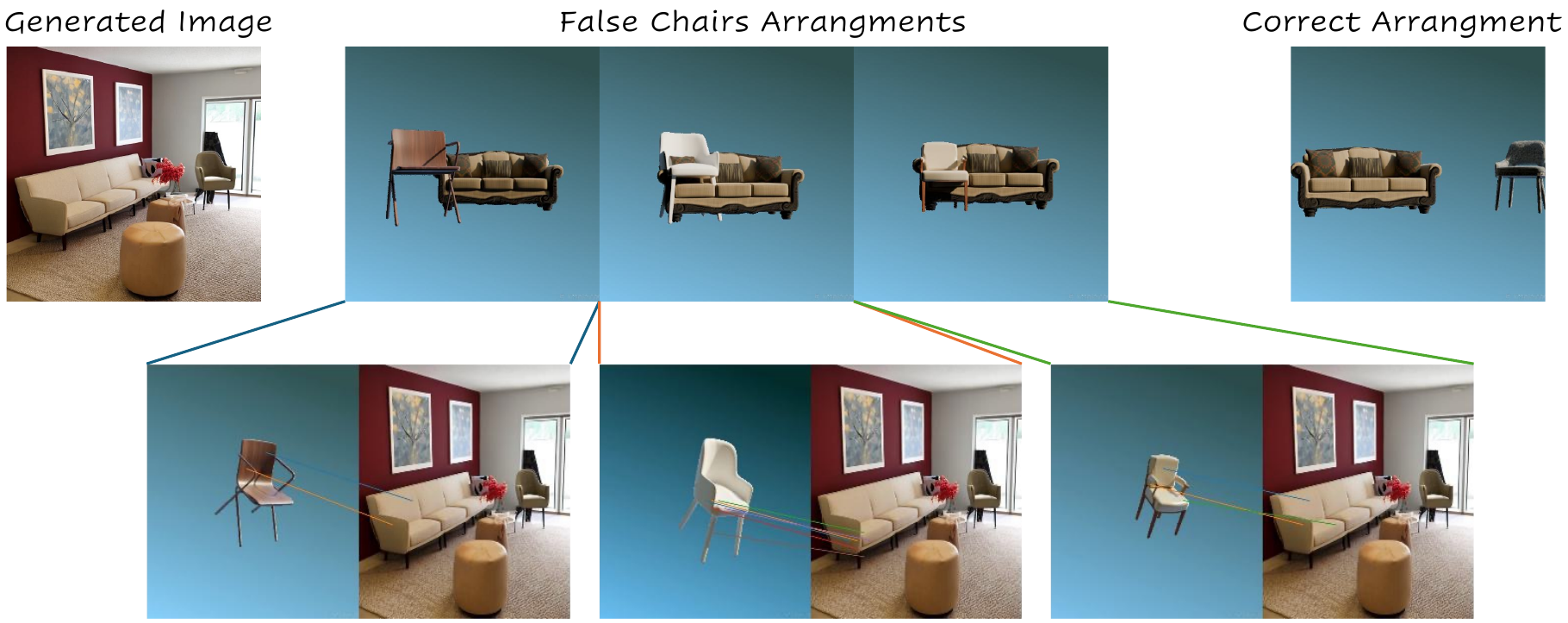}
    \caption{\textbf{Ablation experiment:} Without personalization, the similarity between the input and generated objects affects the matching, and in turn, their placement. The top row shows renders of three output scenes, using three different chairs. If a chair doesn't match the generated one, it might get mixed up with the sofa, as seen in the first three scenes. The rightmost image shows a chair that looks more like the generated one, leading to a better fit. The bottom row highlights the mismatched chairs and where they ended up in the scene. }
    \label{fig:ablation_personalization}
\end{figure}

Without personalization, the appearance or geometry of objects in the generated image was usually different than those of the original object. This has hurt finding corresponding keypoints and the overall performance of the approach.
Figure \ref{fig:ablation_personalization} demonstrates the quality of the matching as a function of appearance similarity.

\subsection{\ourpnp{}}

\begin{figure}[ht]
    \centering
    \includegraphics[width=\textwidth]{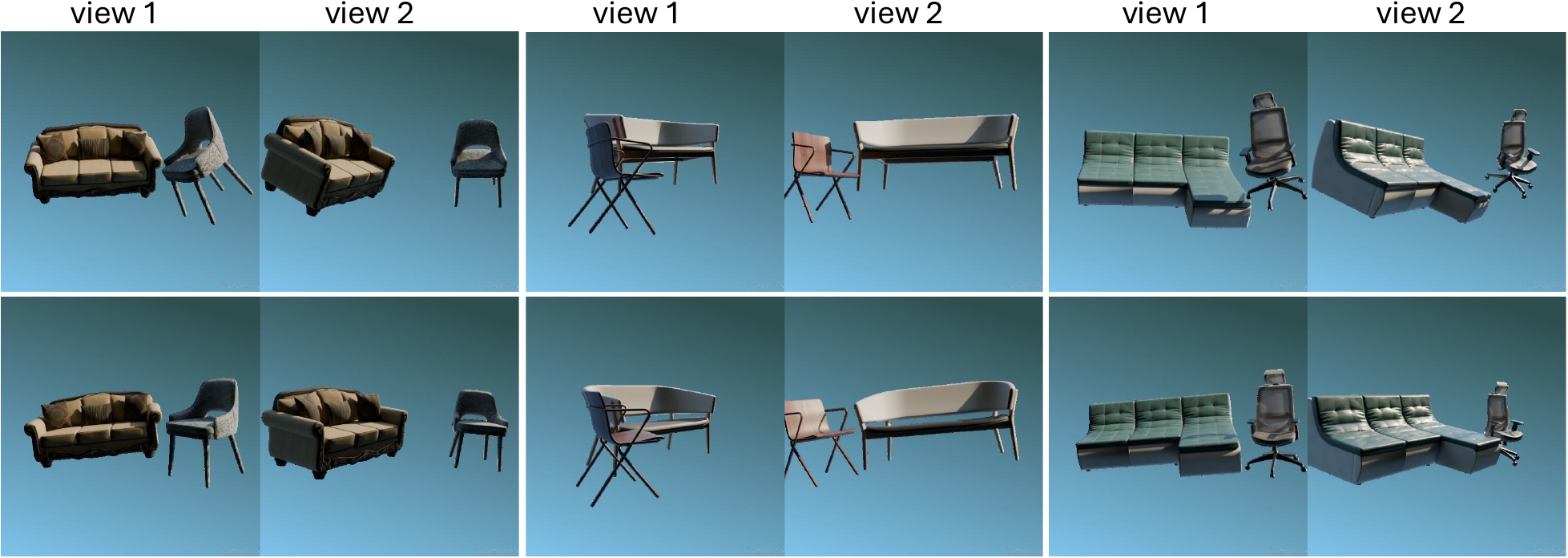}
    \caption{\textbf{Ablation experiment:} The effect of \ourpnp. A qualitative showcase of the joint surface applied by our \ourpnp{} method. The figure is organized into two rows, each presenting a different method. The first presents regular PnP transformations, and the second row presents \ourpnp{} transformations. The different columns present different scenes.}
    \label{fig:ablation_SIPnP}
\end{figure}

We present various viewpoints of the same scene in Fig. \ref{fig:ablation_SIPnP}. The first row depicts object projections using regular PnP, while the second row showcases our \ourpnp{} method, where physical constraints on a joint surface are applied. While regular PnP results in each object lying on a different surface, we achieve improved projection where both objects lay on the same surface with \ourpnp.

\begin{figure}[ht]
    \includegraphics[width=\textwidth]{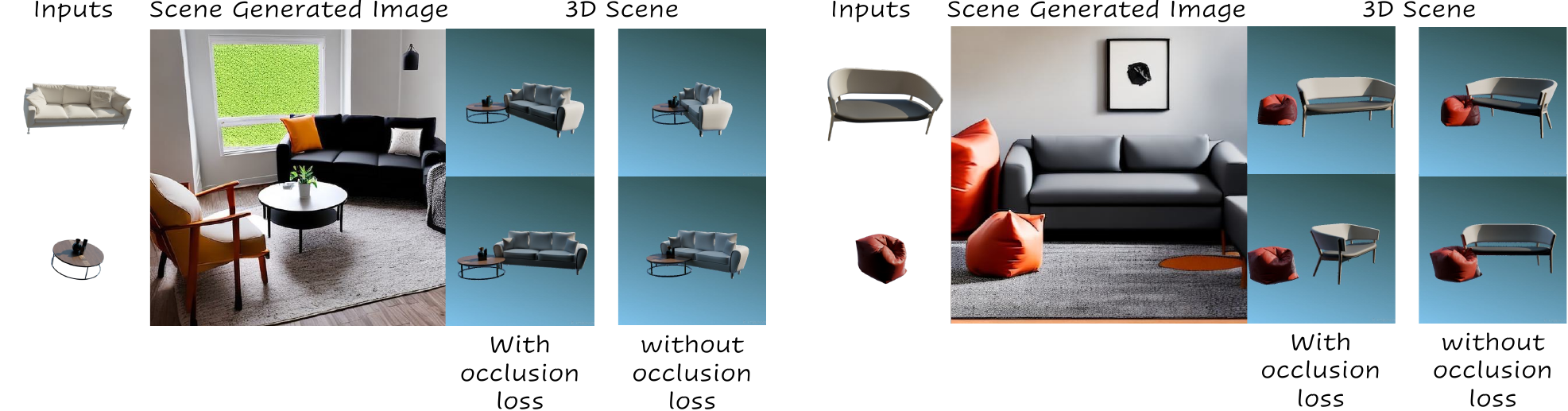}
    \caption{The collision loss prevents object overlap. The figure shows two scenes, each with two objects, with the generated scene image. Adjacent are two output scenes: one with collision loss applied and one without. These examples illustrate how the collision loss adjusts object positions to avoid overlap. }
    \label{fig:collision}
\end{figure}

Moreover, we showcase two examples of collision results and the appropriate solution we provide with the collision loss term in Fig. \ref{fig:collision}.

\section{Results}
\subsection{Evaluation Metrics} \label{Evaluation Metrics}
\begin{figure}[ht]
    \centering
      \includegraphics[width=0.9\textwidth]{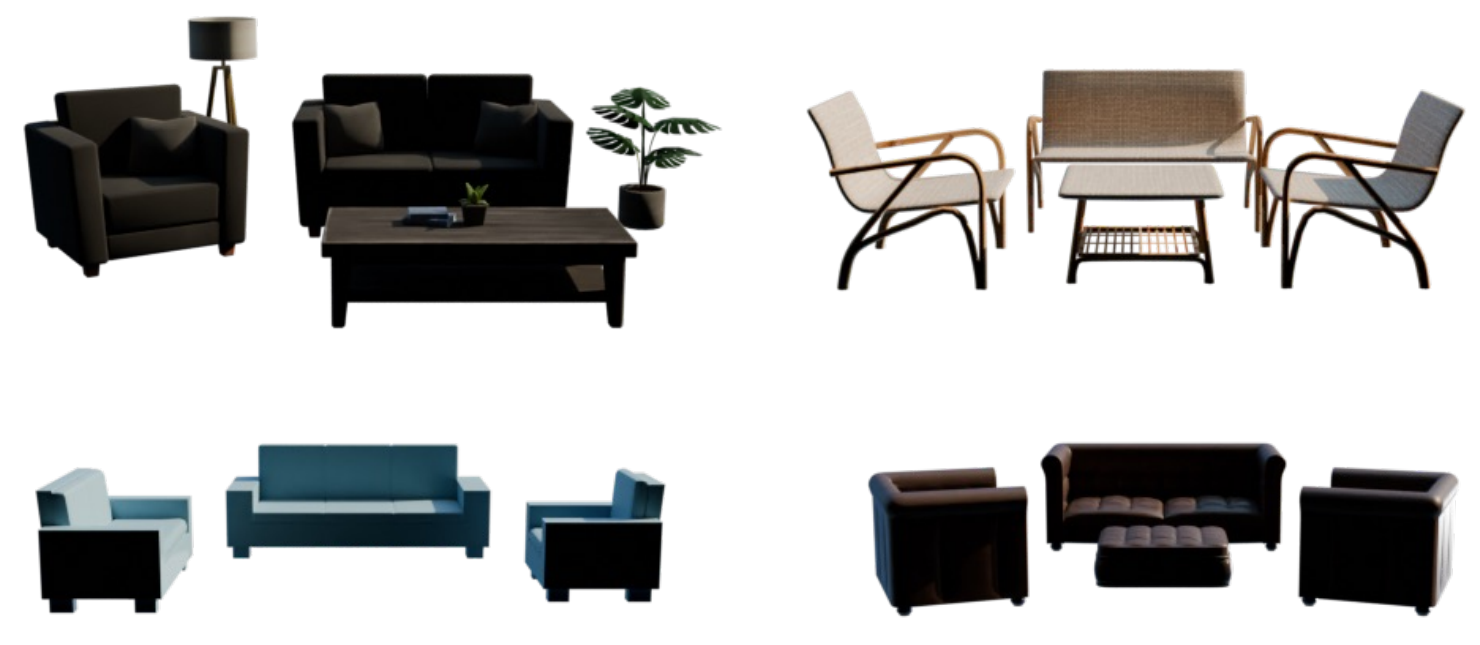}
    \caption{objects of "sofa set" from the Objaverse dataset. } 
    \label{fig:objavers_sofa}
\end{figure}

We provide more details about our quantitative measures.
For each scene category, 10 object scenes were selected from the Objaverse dataset shown in Figure \ref{fig:objavers_sofa}. Since objects in Objaverse have been arranged manually by humans, they serve as a representative sample of potential human arrangements of objects within a scene. The results are presented in Table~\ref{tab:fid_kid_comparison}.
Clip similarity scores between the scene 2D renders and the scene description are shown in Table~\ref{tab:clipscore_comparison}.

\paragraph{Fréchet Inception Distance (FID)} \cite{heusel2017gans}: FID measures the similarity between distributions of generated scenes and real scenes (human-arranged scenes from the Objaverse\cite{objaverse}).  It uses feature vectors from a pre-trained Inception network. Lower FID scores indicate greater similarity, suggesting higher fidelity.

\paragraph{Kernel Inception Distance (KID)} \cite{binkowski2018demystifying}: Similar to FID, KID measures the distance between distributions of generated and real scenes. However, KID uses KL divergence and its expected value does not depend on sample size.

\paragraph{CLIP Similarity}: This metric assesses the semantic similarity between generated scenes and their textual descriptions, using the CLIP model. Higher CLIP scores indicate better alignment between scenes and descriptions.

To calculate these three scores, we rendered 2D images of both generated scenes and human-arranged scenes from the Objaverse dataset \cite{objaverse}, ensuring consistent rendering settings(e.g., background, lighting). We then computed FID and KID scores using these images.

\subsection{Iterative Approach}
\begin{figure}[h]
    \centering
    \includegraphics[width=0.88\textwidth]{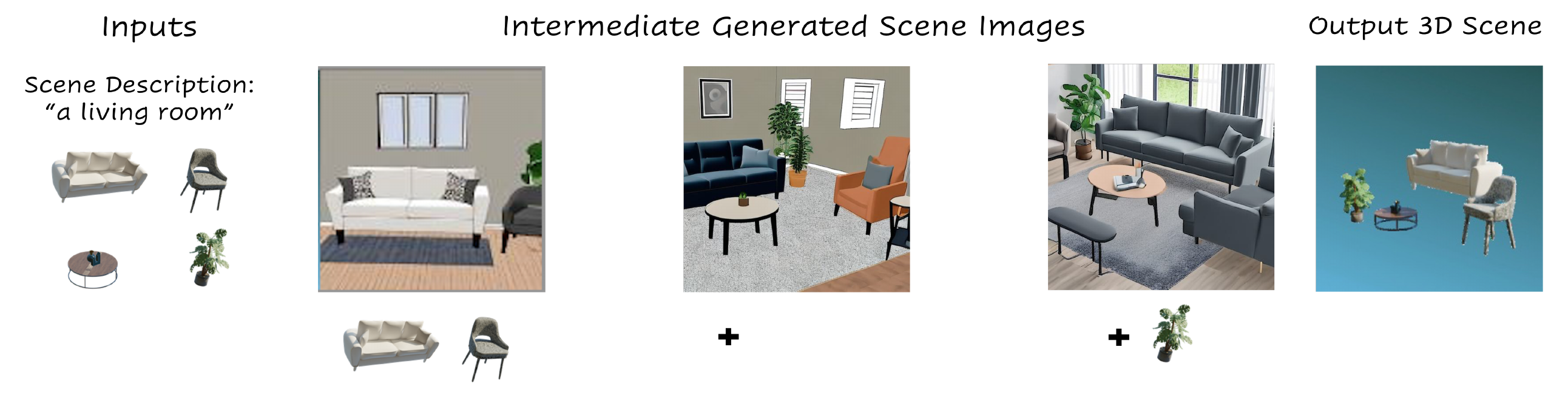}
    \includegraphics[width=0.88\textwidth]{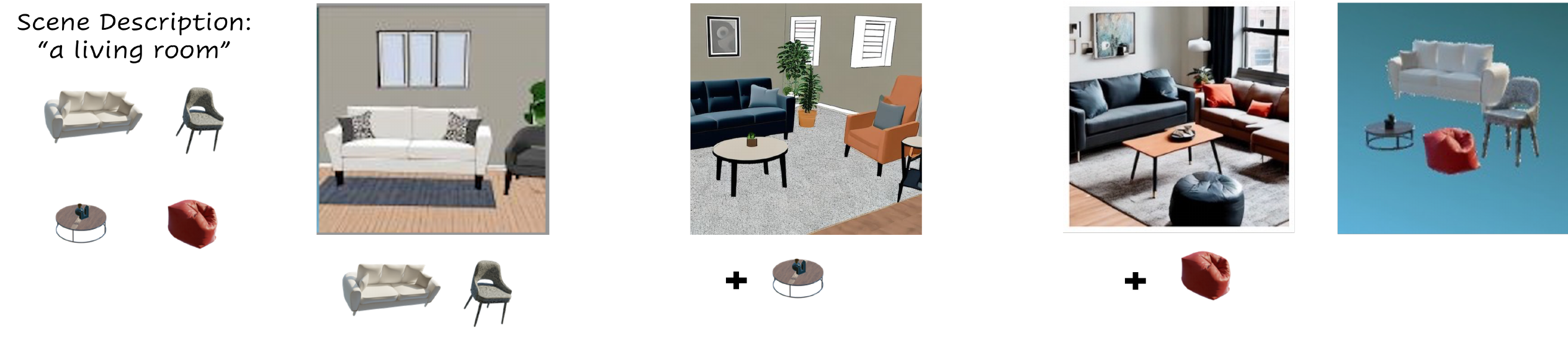}
    \includegraphics[width=0.88\textwidth]{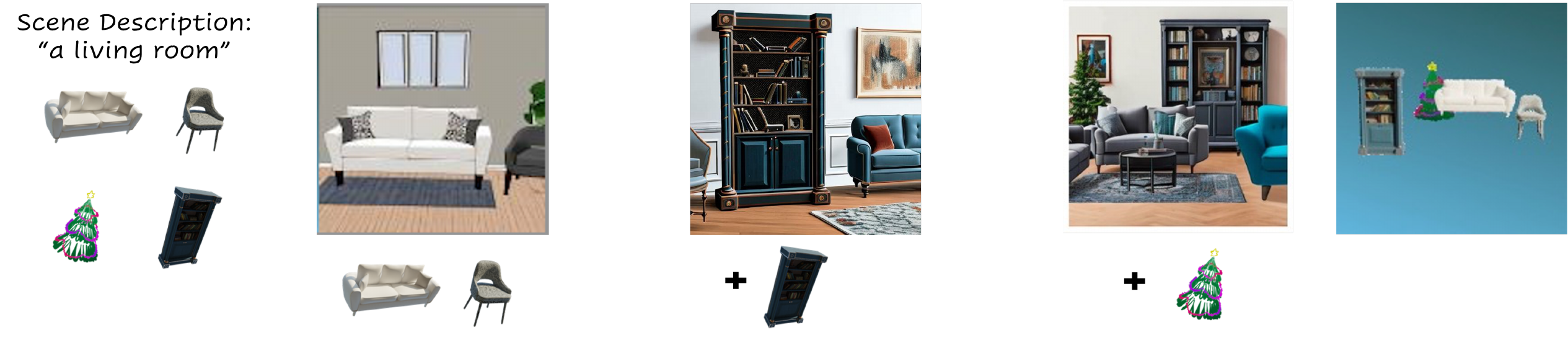}
    \includegraphics[width=0.88\textwidth]{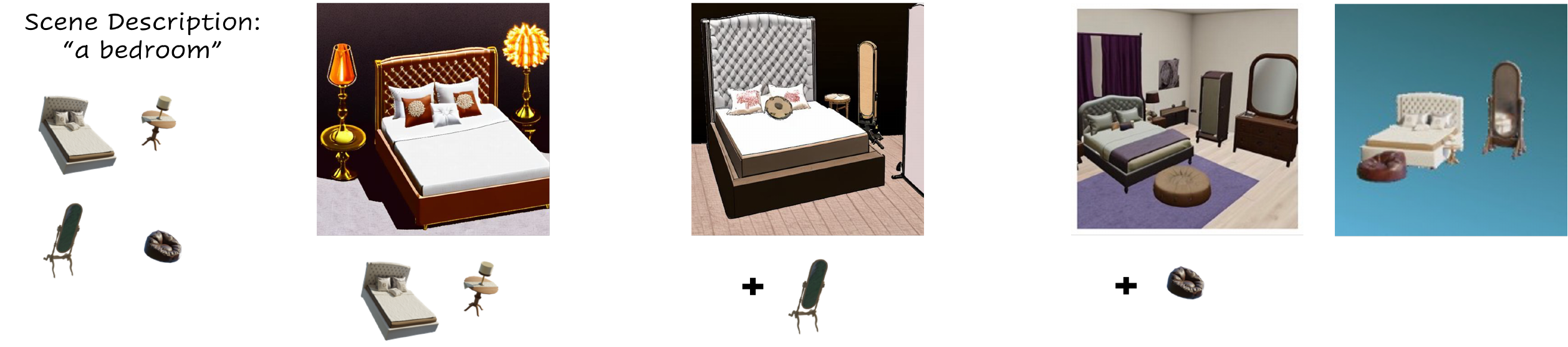}
    \includegraphics[width=0.88\textwidth]{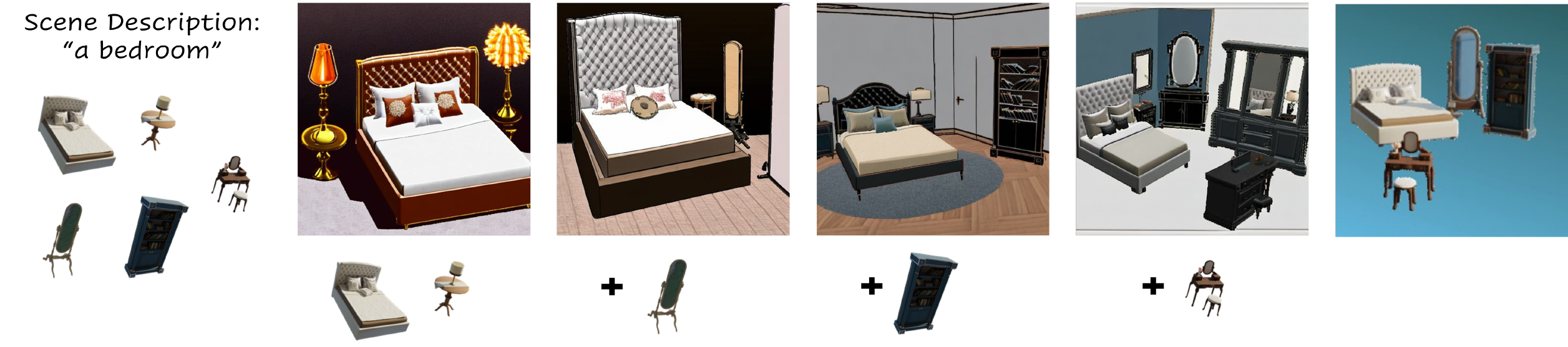}
    \caption{Iterative approach: illustration of the iterative approach results for scenes with four and five objects. We start with two objects and then, in each iteration, we add a new object. For example, in the first row, we started with a scene containing a sofa and a chair, in the second iteration we added a coffee table and in the last iteration, we added a plant.}
    \label{fig:Iterative_Approach_full_4}
\end{figure}

We offer illustrations of our results using the iterative approach, with all the intermediate iterations in figures \ref{fig:Iterative_Approach_full_4}. 

\subsection{Wins and Losses}
\begin{figure}[t]
    \centering
    \includegraphics[width=0.6\textwidth]{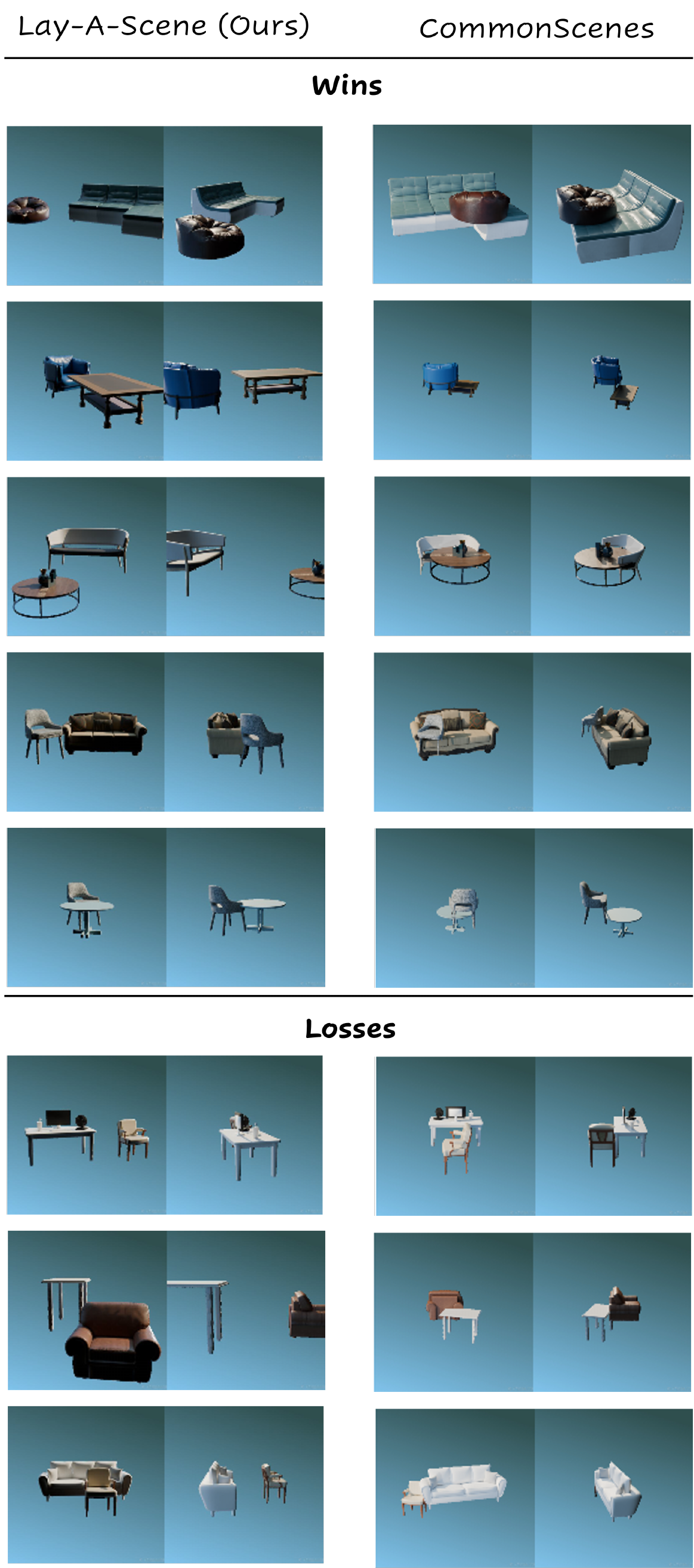}
    \caption{Results Comparison: \ourmethod{} VS CommonScenes. ``wins'' are cases where layouts generated by \ourmethod  were preferred by human raters over layouts generated by CommonScenes.}
    \label{fig:wins_and_losses}
\end{figure}

We demonstrate the success of \ourmethod{} results compared to those of CommonScenes. Figure \ref{fig:wins_and_losses} showcases the wins and losses of \ourmethod{} based on human-rated decisions.

\clearpage

\end{document}